\documentclass{article}

\usepackage[preprint, nonatbib]{neurips_2025}

\usepackage[utf8]{inputenc} % allow utf-8 input
\usepackage[T1]{fontenc}    % use 8-bit T1 fonts
\usepackage{hyperref}       % hyperlinks
\usepackage{url}            % simple URL typesetting
\usepackage{booktabs}       % professional-quality tables
\usepackage{amsfonts}       % blackboard math symbols
\usepackage{nicefrac}       % compact symbols for 1/2, etc.
\usepackage{microtype}      % microtypography
\usepackage{xcolor}         % colors

\usepackage{graphicx}
\usepackage{multirow}
\usepackage{makecell}
\usepackage{amsmath}
\usepackage{subcaption}
\usepackage{pgfplots}
\usepackage{ulem}
\usepackage{filecontents}
\usepackage[numbers]{natbib}
\usepackage[table]{xcolor}

\pgfplotsset{compat=1.18}
\usepgfplotslibrary{statistics}
\usetikzlibrary{calc,positioning,arrows.meta,shapes.geometric}

\title{3-Model Speculative Decoding}

\author{%
 \textbf{Sanghyun Byun},
 \textbf{Mohanad Odema},
 \textbf{Jung Guack},
\\
 \textbf{Baisub Lee},
 \textbf{Jacob Song},
 \textbf{Woo Seong Chung}
\\
\\
LG Electronics USA
}

\begin{document}

\maketitle

% PyramidSD inserts an intermediate qualifier model to bridge the distributional gap, further accelerating LLM inference.

\begin{abstract}
Speculative decoding (SD) accelerates inference in large language models (LLMs) by using a smaller draft model to propose tokens, which are then verified by a larger target model. However, the throughput gains of SD are fundamentally limited by a trade-off between draft model size and token acceptance: smaller draft models generate tokens more quickly but exhibit greater divergence from the target model, resulting in lower acceptance rates and reduced speedups. We introduce \textit{Pyramid Speculative Decoding (PyramidSD)}, an extension of SD that inserts an intermediate qualifier model between the draft and target to bridge the distributional gap in output predictions, allowing smaller model to be used for drafting. This hierarchical decoding strategy improves alignment across models, enabling higher acceptance rates and allowing the use of significantly smaller draft models without sacrificing overall performance. PyramidSD builds on fuzzy acceptance criteria to support relaxed divergence thresholds at each stage, improving throughput. In experiments, PyramidSD achieves up to 1.91× generation speed over standard SD, reaching 124 tokens per second on a consumer GPU (RTX 4090). In small-memory settings with a 1B-parameter draft model and an 8B target model, PyramidSD minimally trades target model quality for improved throughput. Overall, PyramidSD offers a practical approach to enhancing speculative decoding efficiency and can be readily applied to existing inference pipelines.
\end{abstract}

\section{Introduction}

Large language models (LLMs) have achieved remarkable performance across a wide range of natural language processing tasks. However, their rapidly increasing parameter count has led to prohibitively high inference costs, making fast decoding a critical challenge for real-world deployment. Generating each token with a large model result in high latency (e.g., approximately 100ms for Llama-3.2-70B with IQ2\_S on an RTX 4090), limiting interactivity and scalability.

To address these limitations, \textit{speculative decoding} (SD)~\cite{spec-decoding,fast-inference} has emerged, leveraging a smaller, faster draft model to propose multiple tokens, which are then verified by a larger target model. If the target agrees with the draft’s predictions, the tokens are accepted; otherwise, decoding falls back to the target. The overall throughput improves when the acceptance rate is high. However, as the size gap between the draft and target models increases, their output distributions diverge as well, leading to lower acceptance rates and diminishing the speedups achievable with SD.

We propose \textit{Pyramid Speculative Decoding (PyramidSD)}, a new decoding framework that introduces an intermediate qualifier model between the draft and target models. This three-model hierarchy performs incremental verification across two speculative stages, bridging the distributional gap and improving acceptance rates without compromising correctness. PyramidSD builds on fuzzy speculative decoding by applying relaxed divergence thresholds, enabling higher throughput with smaller draft models. PyramidSD requires no additional training and works seamlessly with off-the-shelf families of models. Our key contributions are as follows:

\begin{itemize}
    \item We introduce a novel three-model decoding framework that accelerates inference with an intermediate qualifier model by layering two speculative steps.
    \item We derive two extensions of the proposed speculative decoding criterion, allowing fine-grained trade-off between performance, stability, and speed.
    \item We experimentally demonstrate that PyramidSD achieves higher decoding speeds than standard SD while maintaining relative performance.
    \item We provide a quantitative analysis of PyramidSD’s acceleration behavior, showing how speculative lengths and fuzzy thresholds interact to determine efficiency.
\end{itemize}

\section{Related Works} 
Several approaches proposed to improve speculative decoding efficiency. \textit{Medusa}~\cite{medusa} and \textit{EAGLE}~\cite{eagle} enhance draft quality through specialized architectures and uncertainty modeling. \textit{SpecInfer}~\cite{specinfer} introduces token tree verification to explore multiple decoding paths simultaneously, while \textit{SpecTr}~\cite{spectr} applies optimal transport theory for improved token matching. \textit{Lookahead decoding}~\cite{lookahead} further accelerates inference by breaking sequential dependencies through parallel speculation windows.

Most relevant to our work is \textit{Fuzzy Speculative Decoding}~\cite{fuzzy-spec-decoding}, which relaxes the strict equality constraint in standard SD by allowing tokens to be accepted when their output distributions are sufficiently similar under a divergence threshold. This approach improves acceptance rates but still faces significant limitations when the draft-target gap is large.

Recent work has also explored multi-stage speculative decoding. \textit{Cascade speculative drafting}~\cite{cascade-speculative} chains multiple draft models, and \textit{staged speculative decoding}~\cite{staged-speculative} introduces progressive verification. However, these methods often require specialized training or complex coordination mechanisms, which limit their practicality. Similarly, \textit{Ouroboros}~\cite{ouroboros} reuses recurring phrases to propose longer drafts with fewer forward passes, offering sizable acceleration without fine-tuning.

This motivates the need for approaches that preserve SD's acceleration benefits while maintaining high acceptance rates even across large performance gaps. Notably, many LLMs are released as families of models (e.g., Llama 3.2/3.1: 1B/3B/8B~\cite{llama3}) that share tokenizers and vocabularies. This architectural compatibility presents an underexplored opportunity: leveraging intermediate-sized models to bridge the distributional gap between small draft models and large target models.

\section{3-Model Speculative Decoding}

\newcommand{\pyrscale}{0.53} % <-- set this to (say) 0.8 to shrink everything
\newcommand{\decscale}{0.53}

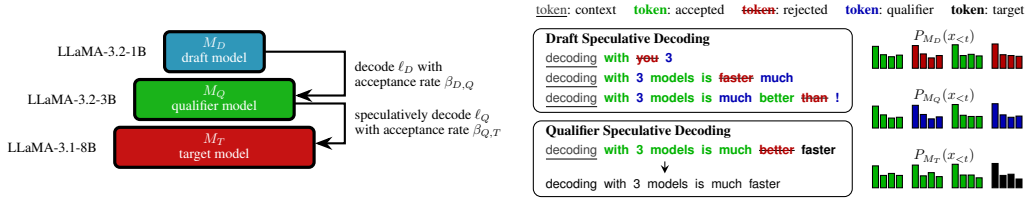
\begin{figure}[t]
\centering
$\vcenter{\hbox{\begin{tikzpicture}[
    scale=\pyrscale,
    every node/.style={transform shape},
    >=Stealth,
    level/.style={draw, rounded corners=2pt, minimum height=10mm, align=center, very thick, text=white},
    lab/.style={anchor=east, align=right},
    arrowline/.style={thick, ->, rounded corners=0pt}
]

% --- geometry ---
\def\h{1.2}          % vertical spacing
\def\wTop{2.5}
\def\wMid{4.0}
\def\wBot{5.0}

% --- pyramid levels with colors ---
\node (top) [level, minimum width=\wTop cm, fill=cyan!60!darkgray] at (0, 2*\h)
    {\begin{tabular}{c}$M_D$\\ draft model\end{tabular}};

\node (mid) [level, minimum width=\wMid cm, fill=green!60!darkgray] at (0, \h)
    {\begin{tabular}{c}$M_Q$\\ qualifier model\end{tabular}};

\node (bot) [level, minimum width=\wBot cm, fill=red!70!darkgray] at (0, 0)
    {\begin{tabular}{c}$M_T$\\ target model\end{tabular}};

% --- left-hand labels ---
\node[lab] at ($(top.west)+(-0.3,0)$) {LLaMA-3.2-1B};
\node[lab] at ($(mid.west)+(-0.3,0)$) {LLaMA-3.2-3B};
\node[lab] at ($(bot.west)+(-0.3,0)$) {LLaMA-3.1-8B};

% --- arrows on the right ---
% Arrow from top to mid
\draw[arrowline] (top.east) -- ++(2,0) -- ++(0,-1.1) -- ($(mid.east)+(0,0.1)$);
\node[anchor=west, align=left] at ($(4,0|-mid.east)+(-0.6,0.6)$)
  {decode $\ell_D$ with\\acceptance rate $\beta_{D,Q}$};

% Arrow from mid to bot
\draw[arrowline] ($(mid.east)-(0,0.1)$) -- ++(1.25,0) -- ++(0,-1) -- ($(bot.east)+(0,0.1)$);
\node[anchor=west, align=left] at ($(4,0|-bot.east)+(-0.6,0.6)$)
  {speculatively decode $\ell_Q$\\with acceptance rate $\beta_{Q,T}$};

\end{tikzpicture}}}$%
\quad
$\vcenter{\hbox{\begin{tikzpicture}[
  scale=\decscale,
  every node/.style={transform shape},
  legendbox/.style={rounded corners=3pt, inner sep=2pt},
  tokenline/.style={font=\sffamily\footnotesize},
  bar/.style={draw, fill=black!60}
]

% --------- Legend (Top Row) ---------
\node (legend) [legendbox, align=left, fill=white] {
  \underline{\textcolor{darkgray}{token}}: context \quad
  \textbf{\textcolor{green!70!black}{token}}: accepted \quad
  \sout{\textbf{\textcolor{red!70!black}{token}}}: rejected \quad
  \textbf{\textcolor{blue!70!black}{token}}: qualifier \quad
  \textbf{\textcolor{black}{token}}: target
};

% --------- Draft Box ---------------
\node[draw, rounded corners=3pt, minimum width=8cm, minimum height=2.1cm, below=0.2cm, anchor=north west] at (legend.south west)(draftbox) {};
\node[below right] at ([xshift=0.2cm,yshift=0cm]draftbox.north west){\textbf{Draft Speculative Decoding}};

% Draft lines
\node[tokenline, anchor=west] at ([xshift=0.2cm,yshift=-0.8cm]draftbox.north west) {
  \underline{\textcolor{darkgray}{decoding}} \ 
  \textbf{\textcolor{green!70!black}{with}} \ 
  \sout{\textbf{\textcolor{red!70!black}{you}}} \
  \textbf{\textcolor{blue!70!black}{3}} \ 
};
\node[tokenline, anchor=west] at ([xshift=0.2cm,yshift=-1.3cm]draftbox.north west) {
  \underline{\textcolor{darkgray}{decoding}} \ 
  \textbf{\textcolor{green!70!black}{with}} \ 
  \textbf{\textcolor{blue!70!black}{3}} \ 
  \textbf{\textcolor{green!70!black}{models}} \ 
  \textbf{\textcolor{green!70!black}{is}} \ 
  \sout{\textbf{\textcolor{red!70!black}{faster}}} \
  \textbf{\textcolor{blue!70!black}{much}} \ 
};
\node[tokenline, anchor=west] at ([xshift=0.2cm,yshift=-1.8cm]draftbox.north west) {
  \underline{\textcolor{darkgray}{decoding}} \ 
  \textbf{\textcolor{green!70!black}{with}} \ 
  \textbf{\textcolor{blue!70!black}{3}} \ 
  \textbf{\textcolor{green!70!black}{models}} \ 
  \textbf{\textcolor{green!70!black}{is}} \ 
  \textbf{\textcolor{blue!70!black}{much}} \ 
  \textbf{\textcolor{green!70!black}{better}} \ 
  \sout{\textbf{\textcolor{red!70!black}{than}}} \
  \textbf{\textcolor{blue!70!black}{!}} \ 
};

% --------- Qualifier Box ---------------
\node[draw, rounded corners=3pt, minimum width=8cm, minimum height=1.9cm, below=0.2cm of draftbox, anchor=north] (qualbox) {};
\node[below right] at ([xshift=0.2cm,yshift=0cm]qualbox.north west){\textbf{Qualifier Speculative Decoding}};

% Qualifier line
\node[tokenline, anchor=west] at ([xshift=0.2cm,yshift=-0.8cm]qualbox.north west) {
  \underline{\textcolor{darkgray}{decoding}} \ 
  \textbf{\textcolor{green!70!black}{with}} \ 
  \textbf{\textcolor{green!70!black}{3}} \ 
  \textbf{\textcolor{green!70!black}{models}} \ 
  \textbf{\textcolor{green!70!black}{is}} \ 
  \textbf{\textcolor{green!70!black}{much}} \ 
  \sout{\textbf{\textcolor{red!70!black}{better}}} \
  \textbf{\textcolor{black}{faster}} \ 
};

\node[tokenline, anchor=west](qualend) at ([xshift=0.2cm,yshift=-1.6cm]qualbox.north west) {
  \textcolor{black}{decoding} \ 
  \textcolor{black}{with} \ 
  \textcolor{black}{3} \ 
  \textcolor{black}{models} \ 
  \textcolor{black}{is} \ 
  \textcolor{black}{much} \ 
  \textcolor{black}{faster} \ 
};

\draw [-stealth]([yshift=0.3cm]qualend.north) -- (qualend.north);

% --------- Bar Charts (right of boxes) ---------

% Draft logits
\foreach \i in {0,...,0} {
  \begin{scope}[shift={([
      xshift={0.5cm+(\i*1.0cm)},
      yshift={-1.0cm}
  ]draftbox.north east)}]
    \draw[bar,fill=green!70!black] (0,0) rectangle (0.15, {0.55+0.02*\i});
    \draw[bar,fill=green!70!black] (0.2,0) rectangle (0.35, {0.32+0.02*mod(\i+1,3)});
    \draw[bar,fill=green!70!black] (0.4,0) rectangle (0.55, {0.35-0.03*mod(\i+2,4)});
    \draw[bar,fill=green!70!black] (0.6,0) rectangle (0.75, {0.33-0.01*\i});
  \end{scope}
}
\foreach \i in {1,...,1} {
  \begin{scope}[shift={([
      xshift={0.5cm+(\i*1.0cm)},
      yshift={-1.0cm}
  ]draftbox.north east)}]
    \draw[bar,fill=red!70!black] (0,0) rectangle (0.15, {0.55+0.02*\i});
    \draw[bar,fill=red!70!black] (0.2,0) rectangle (0.35, {0.32+0.02*mod(\i+1,3)});
    \draw[bar,fill=red!70!black] (0.4,0) rectangle (0.55, {0.35-0.03*mod(\i+2,4)});
    \draw[bar,fill=red!70!black] (0.6,0) rectangle (0.75, {0.33-0.01*\i});
  \end{scope}
}
\foreach \i in {2,...,2} {
  \begin{scope}[shift={([
      xshift={0.5cm+(\i*1.0cm)},
      yshift={-1.0cm}
  ]draftbox.north east)}]
    \draw[bar,fill=green!70!black] (0,0) rectangle (0.15, {0.55+0.02*\i});
    \draw[bar,fill=green!70!black] (0.2,0) rectangle (0.35, {0.32+0.02*mod(\i+1,3)});
    \draw[bar,fill=green!70!black] (0.4,0) rectangle (0.55, {0.35-0.03*mod(\i+2,4)});
    \draw[bar,fill=green!70!black] (0.6,0) rectangle (0.75, {0.33-0.01*\i});
  \end{scope}
}
\foreach \i in {3,...,3} {
  \begin{scope}[shift={([
      xshift={0.5cm+(\i*1.0cm)},
      yshift={-1.0cm}
  ]draftbox.north east)}]
    \draw[bar,fill=red!70!black] (0,0) rectangle (0.15, {0.55+0.02*\i});
    \draw[bar,fill=red!70!black] (0.2,0) rectangle (0.35, {0.32+0.02*mod(\i+1,3)});
    \draw[bar,fill=red!70!black] (0.4,0) rectangle (0.55, {0.35-0.03*mod(\i+2,4)});
    \draw[bar,fill=red!70!black] (0.6,0) rectangle (0.75, {0.33-0.01*\i});
  \end{scope}
}
\node at ([xshift=2.3cm,yshift=-0.2cm]draftbox.north east){$P_{M_D}(x_{<t})$};

% Draft logits
\foreach \i in {0,...,0} {
  \begin{scope}[shift={([
      xshift={0.5cm+(\i*1.0cm)},
      yshift={-2.5cm}
  ]draftbox.north east)}]
    \draw[bar,fill=green!70!black] (0,0) rectangle (0.15, {0.55+0.02*\i});
    \draw[bar,fill=green!70!black] (0.2,0) rectangle (0.35, {0.32+0.01*mod(\i+1,3)});
    \draw[bar,fill=green!70!black] (0.4,0) rectangle (0.55, {0.29-0.02*mod(\i+2,4)});
    \draw[bar,fill=green!70!black] (0.6,0) rectangle (0.75, {0.27+0.01*\i});
  \end{scope}
}
\foreach \i in {1,...,1} {
  \begin{scope}[shift={([
      xshift={0.5cm+(\i*1.0cm)},
      yshift={-2.5cm}
  ]draftbox.north east)}]
    \draw[bar,fill=blue!70!black] (0,0) rectangle (0.15, {0.55+0.02*\i});
    \draw[bar,fill=blue!70!black] (0.2,0) rectangle (0.35, {0.32+0.01*mod(\i+1,3)});
    \draw[bar,fill=blue!70!black] (0.4,0) rectangle (0.55, {0.29-0.02*mod(\i+2,4)});
    \draw[bar,fill=blue!70!black] (0.6,0) rectangle (0.75, {0.27+0.01*\i});
  \end{scope}
}
\foreach \i in {2,...,2} {
  \begin{scope}[shift={([
      xshift={0.5cm+(\i*1.0cm)},
      yshift={-2.5cm}
  ]draftbox.north east)}]
    \draw[bar,fill=green!70!black] (0,0) rectangle (0.15, {0.55+0.02*\i});
    \draw[bar,fill=green!70!black] (0.2,0) rectangle (0.35, {0.32+0.01*mod(\i+1,3)});
    \draw[bar,fill=green!70!black] (0.4,0) rectangle (0.55, {0.29-0.02*mod(\i+2,4)});
    \draw[bar,fill=green!70!black] (0.6,0) rectangle (0.75, {0.27+0.01*\i});
  \end{scope}
}
\foreach \i in {3,...,3} {
  \begin{scope}[shift={([
      xshift={0.5cm+(\i*1.0cm)},
      yshift={-2.5cm}
  ]draftbox.north east)}]
    \draw[bar,fill=blue!70!black] (0,0) rectangle (0.15, {0.55+0.02*\i});
    \draw[bar,fill=blue!70!black] (0.2,0) rectangle (0.35, {0.32+0.01*mod(\i+1,3)});
    \draw[bar,fill=blue!70!black] (0.4,0) rectangle (0.55, {0.29-0.02*mod(\i+2,4)});
    \draw[bar,fill=blue!70!black] (0.6,0) rectangle (0.75, {0.27+0.01*\i});
  \end{scope}
}
\node at ([xshift=2.3cm,yshift=-1.7cm]draftbox.north east){$P_{M_Q}(x_{<t})$};

% Qualifier logits
\foreach \i in {0,...,2} {
  \begin{scope}[shift={([
      xshift={0.5cm+(\i*1.0cm)},
      yshift={-4.0cm}
  ]draftbox.north east)}]
    \draw[bar,fill=green!70!black] (0,0) rectangle (0.15, {0.55+0.02*\i});
    \draw[bar,fill=green!70!black] (0.2,0) rectangle (0.35, {0.32-0.01*mod(\i+1,3)});
    \draw[bar,fill=green!70!black] (0.4,0) rectangle (0.55, {0.32+0.02*mod(\i+2,4)});
    \draw[bar,fill=green!70!black] (0.6,0) rectangle (0.75, {0.31-0.03*\i});
  \end{scope}
}
\foreach \i in {3,...,3} {
  \begin{scope}[shift={([
      xshift={0.5cm+(\i*1.0cm)},
      yshift={-4.0cm}
  ]draftbox.north east)}]
    \draw[bar,fill=black!70!black] (0,0) rectangle (0.15, {0.55+0.02*\i});
    \draw[bar,fill=black!70!black] (0.2,0) rectangle (0.35, {0.32-0.01*mod(\i+1,3)});
    \draw[bar,fill=black!70!black] (0.4,0) rectangle (0.55, {0.32+0.02*mod(\i+2,4)});
    \draw[bar,fill=black!70!black] (0.6,0) rectangle (0.75, {0.31-0.03*\i});
  \end{scope}
}
\node at ([xshift=2.3cm,yshift=-3.2cm]draftbox.north east){$P_{M_T}(x_{<t})$};

\end{tikzpicture}}}$

\caption{\textit{PyramidSD} leverages the common practice of training family of models sharing a common tokenizer to effectively accelerate decoding speed. $\ell_D$ tokens are drafted by $M_D$ and verified by $M_Q$ until $\ell_Q$ tokens are generated, which are passed to $M_T$ for final verification. As fuzzy threshold compares the logits, $\tau=(\tau_Q,\tau_T)$ can be adjusted to trade quality and acceptance rate $\beta$. $P_{M_D}$ is replaced with $P_{M_Q}$ in target comparison for closer distribution to target model. Example shown has been implemented on instruction-tuned LLaMA 3.2 and 3.1 family of models~\cite{llama3}.}

\vskip -1.5mm

\label{fig:pyramidsd}
\end{figure}

To push the performance limits of speculative decoding (SD), we exploit the fact that most model families come in at least three sizes with a shared tokenizer. We call this three-model variant \textit{Pyramid Speculative Decoding (PyramidSD)} and describe its components below.

\subsection{Motivation}
Similar to SD, PyramidSD's speedup depends on the acceptance rate of tokens proposed by the draft model $M_D$ and ultimately accepted by the target model $M_T$. As the performance gap between $M_T$ and $M_D$ increases, their output distributions diverge, leading to lower acceptance rates:
\[
\beta = P\left(\text{Div}(P_{M_T}(x_{t}), P_{M_D}(x_{t})) \leq \tau \right),
\]
where $P_M$ denotes the logit distribution over the next token $x_t$, $\text{Div}$ is a divergence measure, and $\tau$ is a divergence threshold. In standard SD, the $\text{Div}$ formula between the target and the draft model is given by the difference in their logits while having $\tau$ = 0, meaning the draft’s top prediction must exactly match the target’s.

PyramidSD addresses the low-acceptance rate problem by introducing an intermediate qualifier model $M_Q$ between $M_D$ and $M_T$, where parameter size $\|theta\|$ follows $\|theta_T\|>\|theta_Q\|>\|theta_D\|$. This is enabled by the prevalence of modern LLM families that share tokenizers and vocabularies, ensuring architectural compatibility across sizes. Empirically, we observe a systematic entropy gradient across model scales (Figure~\ref{fig:analysis-behavior}): larger models produce low-entropy, high-confidence distributions, whereas smaller models have higher uncertainty. The qualifier $M_Q$ sits naturally between these extremes, making it well-suited to refine the draft’s predictions by filtering out low-confidence tokens before they reach the target. This synergistically offers speedups to the costly inference process.

\subsection{PyramidSD}
If we directly extend the standard SD token acceptance criterion to three models, we encounter a key limitation: the qualifier $M_Q$ contributes no benefit. Under strict equality matching (assisted decoding), a token $x_t$ would be accepted only if the following holds:
\[
P_{M_T}(x_t) \geq P_{M_Q}(x_t) \geq P_{M_D}(x_t), 
\quad x_t = \arg\max_{x \in \nu} P_{M_D}(x),
\]
where $\nu$ is the vocabulary. Since the draft must already agree with the target, $M_Q$ becomes redundant, nullifying its role. To overcome this, PyramidSD adopts Fuzzy Speculative Decoding~\cite{fuzzy-spec-decoding} (FSD), relaxing acceptance criteria based on divergence thresholds instead of strict equality:
\[
\text{Div}(P_{M_Q}(x_t), P_{M_D}(x_t)) \leq \tau_Q
\quad \cap \quad
\text{Div}(P_{M_T}(x_t), P_{M_Q}(x_t)) \leq \tau_T.
\]
This two-stage criterion allows the qualifier to approve plausible tokens from the draft, increasing the effective acceptance rate $\beta$, while the target verifies correctness with more relaxed thresholds. Crucially, PyramidSD assumes $P_{M_Q}$ lies closer to $P_{M_T}$ than $P_{M_D}$, enabling efficient bridging between distributions.

\subsection{Acceleration Analysis}
The throughput (tokens/sec) of speculative decoding $\tilde V_{SD}$ with speculative length $\ell_D$ depends on both the acceptance rate $\beta_{T,D}$ and model speeds $\tilde V_D, \tilde V_T$. Following~\citet{spec-decoding}, the throughput under SD is:
\[
\tilde V_{SD} = \beta_{T,D}(\ell_D + 1) \Bigg/ \left( \frac{\ell_D}{\tilde V_D} + \frac{1}{\tilde V_T} \right).
\]
This extends naturally to FSD, where $\beta_{T,D}$ reflects divergence-threshold-based acceptance rather than exact token matching.

For PyramidSD, acceleration arises from two nested speculative stages: draft $\to$ qualifier and qualifier $\to$ target. We first compute the effective speed of fuzzy speculation between $M_D$ and $M_Q$:
\[
\tilde V_{FSD_{Q,D}} = \beta_{Q,D}(\ell_D + 1) \Bigg/ \left( \frac{\ell_D}{\tilde V_D} + \frac{1}{\tilde V_Q} \right).
\]
Then the overall PyramidSD throughput becomes:
\[
\tilde V_{PSD} = \beta_{T,Q}(\ell_Q + 1) \Bigg/ \left( \frac{\ell_Q}{\tilde V_{FSD_{Q,D}}} + \frac{1}{\tilde V_T} \right).
\]
This decomposition highlights two key design levers: speculative lengths $(\ell_D,\ell_Q)$ and acceptance thresholds $(\tau_Q, \tau_T)$. Larger speculative lengths improve parallelism but may also reduce acceptance due to error compounding, while tighter thresholds stabilize acceptance at the cost of speed. Achieving optimal acceleration thus requires balancing these competing effects.

\subsection{Assisted Decoding Variant}
We introduce an assisted variant of PyramidSD (PSD$_A$) which improves stability by leveraging assisted decoding in place of $\tau_Q$. When draft tokens are rejected by $M_Q$, instead of using stnadard SD probability rule, PSD$_A$ samples directly from the qualifier’s distribution. This provides a guaranteed quality floor matching $M_Q$ while still accelerating decoding via $M_D$, assuming no significant acceptance rate drop exists. 

This design is motivated by the trade-off between speed and quality. High $\tau_Q$ values increase throughput by allowing more tokens from $M_D$ to pass, but at the risk of forwarding low-quality predictions to $M_T$, potentially degrading output quality. PSD$_A$ resolves this tension by using $M_Q$ as a fallback generator, combining the draft’s acceleration with the qualifier’s stability at the cost of reduced acceleration.

\section{Experiments}

\pgfdeclareplotmark{mystar}{
    \node[star,star point ratio=2.25,minimum size=13pt,
          inner sep=0pt,draw=black,solid,fill=yellow] {};
}

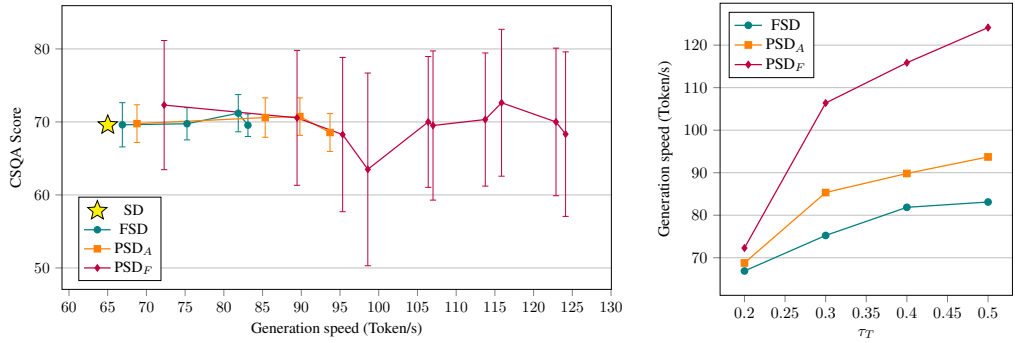
\begin{figure}[t]
\centering
$\vcenter{\hbox{\resizebox{0.6\linewidth}{!}{
\begin{tikzpicture}
\begin{axis}[
    xlabel={Generation speed (Token/s)},
    ylabel={CSQA Score},
    legend pos=south west,
    ymajorgrids,
    width=14cm,
    height=8cm,
    axis lines=box,
    tick align=outside,
    xtick pos=bottom,
    ytick pos=left,
    mark options={scale=1},
]

\addplot[ only marks, mark=mystar, ] coordinates {(65.01, 69.58)};
\addlegendentry{SD}

\addplot [color=teal, thick, mark=*,]
 plot [error bars/.cd, y dir = both, y explicit]
 table[y error index=2]{fsd.dat};
\addlegendentry{FSD}

\addplot [color=orange, thick, mark=square*,]
 plot [error bars/.cd, y dir = both, y explicit]
 table[y error index=2]{psd-a.dat};
\addlegendentry{PSD$_A$}

\addplot [color=purple, thick, mark=diamond*,]
 plot [error bars/.cd, y dir = both, y explicit]
 table[y error index=2]{psd-f.dat};
\addlegendentry{PSD$_F$}

\end{axis}
\end{tikzpicture}
}}}$%
\hspace{4pt}
$\vcenter{\hbox{\resizebox{0.35\linewidth}{!}{
\begin{tikzpicture}
\begin{axis}[
    xlabel={$\tau_T$},
    ylabel={Generation speed (Token/s)},
    legend pos=north west,
    ymajorgrids,
    width=8cm,
    height=8cm,
    axis lines=box,
    tick align=outside,
    xtick pos=bottom,
    ytick pos=left,
    mark options={scale=1},
]

\addplot [color=teal, thick, mark=*,] coordinates {
(0.2, 66.89)
(0.3, 75.24)
(0.4, 81.86)
(0.5, 83.11)
};
\addlegendentry{FSD}

\addplot [color=orange, thick, mark=square*,] coordinates {
(0.2, 68.78)
(0.3, 85.34)
(0.4, 89.82)
(0.5, 93.71)
};
\addlegendentry{PSD$_A$}

\addplot [color=purple, thick, mark=diamond*,] coordinates {
(0.2, 72.29)
(0.3, 106.39)
(0.4, 115.86)
(0.5, 124.13)
};
\addlegendentry{PSD$_F$}

\end{axis}
\end{tikzpicture}
}}}$%
\caption{(Right) Evaluation result comparison on CSQA~\cite{talmor-etal-2019-commonsenseqa} dataset with Llama-3.1-8B, Llama-3.2-3B, and Llama-3.2-1B~\cite{llama3} Generation speeds for FSD, PSD$_A$, and PSD$_F$ are majorly controlled with $\tau_T$ (FSD, PSD) and $\tau_Q$ (PSD) adjustments. We report the fastest generation speed for each data point. Error bars represent standard deviation for CSQA score. (Left) Generation speed comparison with varying $\tau_T$. Highest speeds are reported.}

\vskip -1.5mm

\label{fig:result-eval}
\end{figure}

% \paragraph{Experimental Setup}

% We evaluate PyramidSD on diverse language generation tasks using the Llama 3.2 and 3.1 model families~\cite{llama3}. Our experiments are conducted on 2× NVIDIA RTX 4090 GPUs with 24GB memory each. We use the Hugging Face Transformers implementation with instruction-tuned variants: Llama 3.2 1B (draft), 3B (qualifier), and Llama 3.1 8B (target). All experiments use a maximum generation length of 2048 tokens and temperature 0.7 for sampling. Following prior work~\cite{fuzzy-spec-decoding}, we evaluate on 5 random samples per prompt and report mean and standard deviation.

% We compare four decoding strategies: (1) Standard Speculative Decoding (SD), (2) Fuzzy Speculative Decoding (FSD), (3) PyramidSD with Assisted decoding (PSD$_A$), and (4) PyramidSD with Fuzzy acceptance (PSD$_F$). For hyperparameter selection, we perform grid search over divergence thresholds $\tau_Q, \tau_T \in \{0, 0.01, 0.05, 0.1\}$ and speculative lengths $\ell_D \in \{2, 4, 6\}$, $\ell_Q \in \{4, 8, 12\}$.

\subsection{Experimental Setup}

We evaluate PyramidSD on CSQA~\cite{talmor-etal-2019-commonsenseqa} evaluation set using the Llama 3.2 and 3.1 model family~\cite{llama3}. All experiments are conducted on RTX 4090 GPUs, each with 24GB of memory. We build on the Hugging Face Transformers~\cite{wolf-etal-2020-transformers} implementation and use instruction-tuned variants: Llama 3.2 1B (draft), Llama 3.2 3B (qualifier), and Llama 3.1 8B (target)~\cite{llama3}. Unless otherwise noted, we set the maximum generation length to 2048 tokens and use a sampling temperature of 0.7. Following prior work~\cite{fuzzy-spec-decoding}, we evaluate each prompt using five pre-determined QA samples and report the mean and standard deviation of results.

We compare four decoding strategies: (1) Standard Speculative Decoding (SD)~\cite{spec-decoding}, (2) Fuzzy Speculative Decoding (FSD)~\cite{fuzzy-spec-decoding}, (3) PyramidSD with Assisted Decoding (PSD$_A$), and (4) PyramidSD with Fuzzy Acceptance (PSD$_F$). For hyperparameter selection, we perform a grid search over divergence thresholds $\tau_Q, \tau_T \in \{0.2, 0.3, 0.4, 0.5\}$ and speculative lengths $\ell_D \in \{2, 4, 6\}$, $\ell_Q \in \{1, 2, 3, 4, 5, 7, 10, 15, 20, 25\}$. This setup allows us to isolate the impact of different parameter configurations on both throughput and stability.

% \paragraph{PyramidSD Performance}

% Figure~\ref{fig:result-eval} demonstrates the superior performance of PyramidSD variants compared to baseline methods. Both PSD$_A$ and PSD$_F$ achieve substantial speedups over standard SD and FSD across varying divergence thresholds. Specifically, PSD$_A$ provides up to 1.44x improvement over SD with minimal variance, making it suitable for production deployments where predictability is crucial.

% PSD$_F$ achieves even higher acceleration (up to 1.91x over SD) but exhibits a larger standard deviation due to the compound effect of stacked divergence thresholds. The increased variability stems from the multiplicative nature of acceptance rates across two speculative stages. However, this aggressive approach yields the best peak performance when properly tuned, as the relaxed acceptance criteria at both stages significantly boost overall throughput.

% Notably, FSD shows faster growth rate with increased threshold $\tau_T$, as higher thresholds allow more aggressive speculation. In PyramidSD, setting $\tau_Q \leq \tau_T$ enables balanced acceleration—the qualifier filters obvious mismatches while maintaining sufficient acceptance for downstream verification. This hierarchical filtering prevents quality degradation while maximizing speed.

\subsection{PyramidSD Performance}

Figure~\ref{fig:result-eval} demonstrates the superior performance of PyramidSD variants compared to baseline methods across varying divergence thresholds. Both PSD$_A$ and PSD$_F$ achieve substantial speedups over standard SD and FSD, confirming the benefits of introducing an intermediate qualifier model. Among the two, PSD$_A$ consistently provides up to 1.44$\times$ improvement over SD while maintaining low variance across runs, making it particularly well-suited for production scenarios where predictable performance is critical.

In contrast, PSD$_F$ delivers even greater acceleration—up to 1.91$\times$ faster than SD—but exhibits a larger standard deviation due to the compound effect of fuzzy acceptance across two speculative stages. The increased variability arises from the multiplicative nature of acceptance rates: errors at the draft-to-qualifier stage can propagate downstream, amplifying output fluctuations. However, when properly tuned, PSD$_F$ achieves the highest peak performance, as relaxed acceptance thresholds at both stages allow significantly larger speculative jumps.

In PyramidSD, performance is best maintained when setting $\tau_Q \leq \tau_T$: the qualifier first filters out clear mismatches while passing plausible candidates forward, allowing the target model to perform final verification efficiently. This hierarchical filtering helps PyramidSD balance speed and quality, achieving further acceleration while mostly preserving target-level correctness.

The benefits of PyramidSD become clearer when examining the performance of the two proposed variants, PSD$_A$ and PSD$_F$. PSD$_A$, which incorporates an assisted fallback, demonstrates robust stability across a wide range of threshold combinations. Its performance closely tracks FSD while achieving higher aggregate speedups and exhibits similarly robust sensitivity to thresholds as FSD. The fallback mechanism allows PyramidSD to reject low-confidence draft outputs early, preventing quality degradation. In contrast, PSD$_F$, which applies fuzzy relaxation at both the draft-to-qualifier and qualifier-to-target stages, achieves higher average throughput but displays noticeably larger variance across runs. This instability likely arises from compounding effects in the two fuzzy stages, where errors introduced early by the draft stage are amplified. These results underscore the importance of carefully tuning the pair $(\tau_Q, \tau_T)$: a permissive $\tau_T$ combined with a moderately strict $\tau_Q$ can maximize PyramidSD’s benefits, but extreme configurations in either direction diminish its advantage.

% \paragraph{Analysis of Model Behavior}
% Our analysis of the Llama model family reveals systematic differences in prediction confidence across scales (detailed in Appendix Figure~\ref{fig:analysis-behavior}). Larger models exhibit lower entropy and higher confidence scores, with the 3B qualifier model showing intermediate characteristics that make it ideal for bridging the 1B-8B gap. This entropy gradient validates our pyramid approach: the qualifier naturally corrects the draft's high-entropy predictions toward the target's confident distribution.

\begin{figure}[t]

\centering
\includegraphics[width=0.5\linewidth]{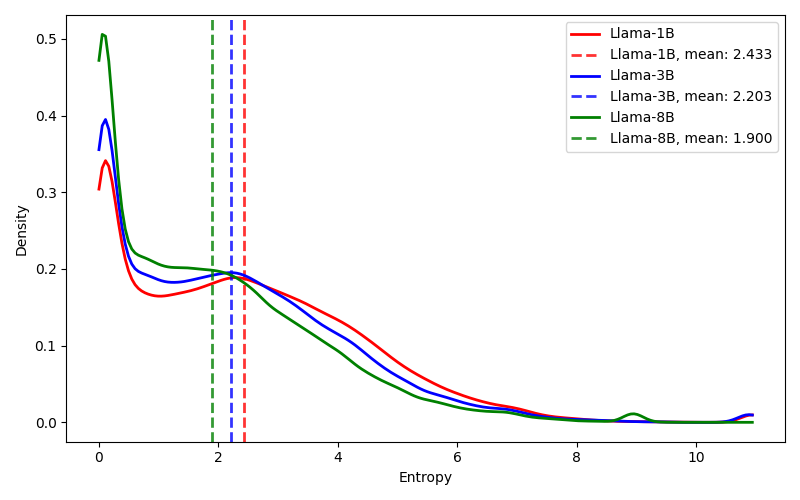}%
\includegraphics[width=0.5\linewidth]{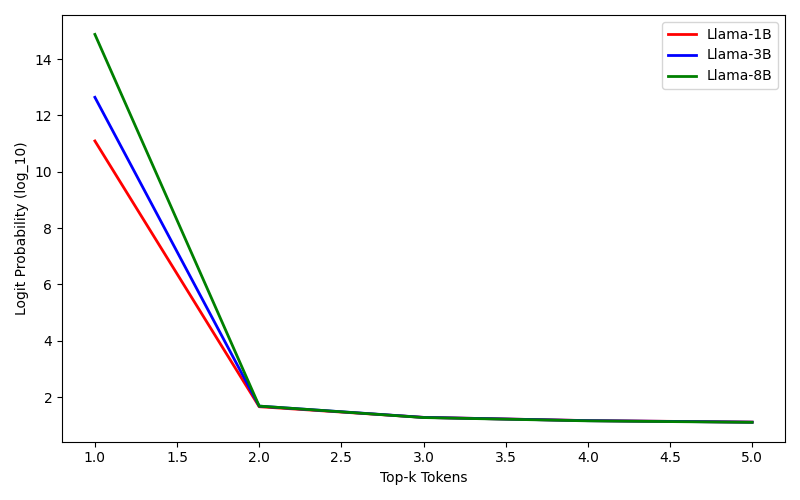}
\caption{Behavior analysis on LLaMA 3.2/3.1 family with 1B, 3B, and 8B models on CSQA dataset. Both entropy and confidence scales with model size, revealing characteristics that can be capitalized for inference speed-up.}

\vskip -1.5mm

\label{fig:analysis-behavior}
\end{figure}

\subsection{Model Behavior Analysis}

% Our analysis of the Llama 3.2/3.1 model family reveals systematic differences in prediction behavior across scales, which provide critical insight into PyramidSD’s effectiveness (Figure~\ref{fig:analysis-behavior}). We evaluate entropy distributions and token confidence scores for models of varying sizes on a diverse set of generation tasks. The results highlight a consistent entropy gradient: larger models exhibit lower mean entropy and produce more confident, peaked output distributions, while smaller models generate higher-entropy predictions, reflecting greater uncertainty. The intermediate-sized 3B model demonstrates entropy characteristics that fall neatly between the 1B draft and 8B target, making it particularly well-suited to act as a qualifier bridging the two extremes.

Figure~\ref{fig:analysis-behavior} illustrates this progression clearly. The 1B draft model produces relatively uniform entropy distributions, suggesting widespread uncertainty across its token predictions. The 3B qualifier model reduces this uncertainty, producing sharper distributions while still retaining flexibility, and the 8B target model shows strongly right-skewed entropy distributions, indicative of highly confident and selective predictions. This smooth gradient supports PyramidSD’s core design: the qualifier operates naturally in an intermediate regime, capable of validating high-quality draft tokens while rejecting low-confidence outputs before they reach the target stage. As a result, the qualifier enables higher acceptance rates.

We further analyze token-level confidence by measuring the maximum predicted probability at each decoding step. The results align with the entropy findings: the 1B draft frequently exhibits low peak probabilities, reflecting its uncertainty and tendency toward diverse token sampling. The 3B qualifier achieves intermediate confidence, producing predictions that are sharper and better aligned with the target distribution. Finally, the 8B target produces consistently high-confidence predictions, selecting fewer but more reliable candidate tokens. This hierarchy of token confidence enables PyramidSD’s cascaded filtering mechanism to function effectively. Together, the entropy gradient and confidence hierarchy demonstrate that the natural scaling properties validate the foundation of the proposed decoding scheme.

% \paragraph{Ablation Studies}
% Comprehensive ablation studies on hyperparameter combinations are presented in Appendix Tables~\ref{table:ablation-threshold} and \ref{table:ablation-full}. Key findings include: (1) optimal $\ell_D/\ell_Q$ ratios typically range from 1:2 to 1:3, balancing draft overhead with qualifier efficiency; (2) threshold selection shows non-monotonic effects, with high-range values maintaining performance; (3) the interaction between $\tau_Q$ and $\tau_T$ is critical—overly permissive $\tau_Q$ degrades quality while overly strict values eliminate PyramidSD's benefits.

\subsection{Ablation Studies}

\begin{table}[t]
\caption{Evaluation result comparison on CSQA~\cite{talmor-etal-2019-commonsenseqa} dataset with LLaMA-3.2-1B (draft for PSD), LLaMA-3.2-3B (qualifier for PSD / draft), and LlaMA-3.1-8B (target)~\cite{llama3} under different $\tau_Q$/$\tau_T$ configurations, compared to vanilla and fuzzy speculative decoding. PSD$_A$ and PSD$_F$ denote that assisted and fuzzy speculative decoding (FSD) was used for qualifier model, respectively.}

\vskip 1.5mm

\centering
\resizebox{\linewidth}{!}{
\begin{tabular}{c*{9}{c}c}

% \Xhline{2\arrayrulewidth}
\toprule

& \multicolumn{2}{c}{SD~\cite{spec-decoding}} & \multicolumn{4}{c}{FSD~\cite{fuzzy-spec-decoding}} & \multicolumn{4}{c}{PSD$_A$} \\
\cmidrule(lr){2-3} \cmidrule(lr){4-7} \cmidrule(lr){8-11}
$\tau_T$
& \multicolumn{2}{c}{-} & 0.2 & 0.3 & 0.4 & 0.5 &
  0.2 & 0.3 & 0.4 & 0.5 \\
%\hline
CSQA & 
\multicolumn{2}{c}{69.58$\pm$2.20} & 
69.61$\pm$3.03 & 69.75$\pm$2.21 & 71.20$\pm$2.55 & 69.55$\pm$1.55 &
69.77$\pm$2.58 & 70.60$\pm$2.70 & 70.73$\pm$2.57 & 68.55$\pm$2.59 \\

\midrule[2pt]

& \multicolumn{10}{c}{PSD$_F$} \\
\cmidrule(lr){1-11}
$\tau_T$ & 
\multicolumn{1}{c}{0.2} & \multicolumn{2}{c}{0.3} & \multicolumn{3}{c}{0.4} & \multicolumn{4}{c}{0.5} \\
%\hline
\cmidrule(lr){2-2} \cmidrule(lr){3-4} \cmidrule(lr){5-7} \cmidrule(lr){8-11}
$\tau_Q$ & 
0.2 & 0.2 & 0.3 & 0.2 & 0.3 & 0.4 & 0.2 & 0.3 & 0.4 & 0.5 \\
%\hline
CSQA & 
72.31$\pm$8.85 & 
70.56$\pm$9.24 & 70.00$\pm$8.96 & 
68.27$\pm$10.57 & 70.33$\pm$9.12 & 72.63$\pm$10.06 & 
63.50$\pm$13.20 & 69.51$\pm$10.22 & 68.33$\pm$11.27 & 70.00$\pm$10.11 \\

%\Xhline{2\arrayrulewidth}
\bottomrule

\end{tabular}}

\vskip -1.5mm

\label{table:ablation-threshold}
\end{table}

Table~\ref{table:ablation-threshold} presents detailed results for different divergence threshold combinations across four decoding strategies. Evaluations are performed on he CommonsenseQA (CSQA)~\cite{talmor-etal-2019-commonsenseqa} becnhmark. We report the evaluation scores with standard deviation across multiple $\ell_D$/$\ell_Q$ pairs. As expected, standard SD maintains a stable CSQS score of $69.58 \pm 2.20$, since it relies on strict equality matching during verification. In contrast, Fuzzy SD exhibits a mild sensitivity to the divergence threshold $\tau_T$. Performance does not scale directly with $\tau_T$, achieving an average of $70.03$ on CSQA; however, we observe that overly permissive thresholds ($\tau_T > 0.5$) degrade output quality without providing proportional speedups. This reflects a core limitation of single-stage fuzzy relaxation: while relaxing the matching criterion can improve acceptance rates, pushing it too far allows misaligned draft tokens to propagate, reducing overall correctness.

We conduct comprehensive ablation studies on divergence thresholds and speculative length configurations, with full results presented in Tables~\ref{table:ablation-full} and \ref{table:ablation-full-2}. We report the mean tokens per second for each configuration. These experiments provide deeper insights into the mechanisms driving PyramidSD's efficiency gains and reveal how the interplay between thresholds, speculative lengths, and verification stages shapes overall throughput.

Larger qualifier lengths (e.g., $\ell_Q = 25$) generally yield higher throughput, as they allow PyramidSD to propose longer speculative windows per iteration. However, simply increasing $\ell_D$ does not uniformly translate into speedups: beyond a certain point, the acceptance rate drops precipitously due to prediction errors. We find that setting $\ell_D$ to approximately half of $\ell_Q$ consistently achieves the best trade-off between speculation depth and acceptance stability, with optimal ratios typically falling between $1:2$ and $1:3$.

Finally, a striking observation from these studies is the presence of non-monotonic performance patterns across both thresholds and speculative lengths. Contrary to intuition, more aggressive speculation—whether through higher thresholds, longer draft sequences, or deeper relaxation—does not always yield higher throughput or lower performance. In several cases, medium-threshold, moderate-length configurations outperform more extreme setups. This suggests that PyramidSD’s efficiency is governed by a subtle trade-off: aggressive speculation accelerates token generation but also increases the likelihood of rejection cascades, which ultimately reduces overall throughput. These findings highlight the need for careful joint tuning of speculative lengths and divergence thresholds to fully utilize the potential of three-model speculation.

\section{Discussion}

PyramidSD introduces additional complexity in its hyperparameter space. The interaction between divergence thresholds $(\tau_Q, \tau_T)$ and speculative lengths $(\ell_D, \ell_Q)$ strongly influences performance, and the optimal configuration varies across tasks, datasets, and model families. This variability further diminishes the plug-and-play capability of speculative decoding. Also, the multiplicative nature of acceptance rates across two speculative stages can lead to performance variability, especially for the fuzzy variant (PSD$_F$), where relaxed thresholds amplify fluctuations in throughput. Finally, PyramidSD currently relies on families of compatible models that share tokenizers and vocabularies, which may limit adoption for architectures without well-aligned models.

Despite these limitations, PyramidSD provides a new perspective on accelerating LLM inference. Our results suggest that hierarchical decoding strategies can exploit intrinsic scaling properties of model families to achieve better trade-offs between speed and accuracy than flat, single-stage approaches. Understanding and formalizing these dynamics opens opportunities for designing more principled speculative decoding algorithms.

\section{Conclusion and Future Work}

We introduced \textit{Pyramid Speculative Decoding (PyramidSD)}, a hierarchical framework that accelerates large language model inference up to 1.91x over vanilla speculative decoding by introducing an intermediate qualifier model between draft and target models. By exploiting the natural entropy gradient across model scales and incorporating fuzzy acceptance criteria, PyramidSD achieves significant speedups while maintaining generation quality.
Looking forward, several directions emerge for extending PyramidSD. One promising avenue is the development of adaptive controllers that dynamically adjust divergence thresholds and speculative lengths based on real-time decoding context, combining PSD$_F$'s peak performance with PSD$_A$'s stability. Another is generalization beyond homogeneous model families: enabling cross-family and mixed-attention setups would broaden PyramidSD’s applicability, especially as emerging architectures diverge in tokenizers and scaling behaviors. Finally, understanding how model size ratios and entropy distributions influence acceptance rates could guide principled model selection and hyperparameter tuning.
Overall, PyramidSD demonstrates that hierarchical speculative decoding is a powerful paradigm for efficient inference. As model families continue to grow and diversify, we believe this approach provides a foundation for next-generation decoding strategies that balance throughput, quality, and scalability.

\bibliographystyle{plainnat}
\bibliography{main}

\appendix

\newpage

\section{Full Ablation Results}

% \input{figs/ablation_threshold}

% Table~\ref{table:ablation-threshold} presents comprehensive ablation studies on divergence threshold combinations for all four decoding strategies. We evaluate on the CommonsenseQA (CSQA) benchmark and report mean tokens/second with standard deviation across 100 samples.

% \paragraph{Key Observations on Thresholds}

% \textbf{Standard SD:} As expected, shows no variation with threshold changes (fixed at 69.58 ± 2.20 tokens/sec) since it uses strict equality matching.

% \textbf{Fuzzy SD:} Performance doesn't seem to scale with $\tau_T$, with average score of 70.03 on CSQA. However, excessive relaxation ($\tau_T > 0.5$) will likely result in quality degradation without proportional speed gains.

% \textbf{PSD$_A$:} Shows robust performance across threshold combinations, without too much variation in performance drops, similarly to fuzzy SD~\cite{fuzzy-spec-decoding}. The assisted fallback mechanism provides stability even with suboptimal thresholds.

% \textbf{PSD$_F$:} Achieves similar average performance across the board but displays much larger deviation, which is likely due to the compound effect of two fuzzy stages amplifying both speedup and instability. Careful tuning of hyperparameters will likely result in optimal speed gains without significant performance drops.

\begin{table}[th]
\caption{Complete generation speed ablation of PSD with fuzzy threshold and draft length exploration. We evaluate the speed on LLaMA-3.2-1B (draft for PSD), LLaMA-3.2-3B (qualifier for PSD / draft), and LlaMA-3.1-8B (target) under different $\tau_T$/$\tau_Q$/$\ell_Q$/$\ell_D$ configurations, compared to vanilla and fuzzy speculative decoding. PSD$_A$ and PSD$_F$ denote that assisted and fuzzy speculative decoding (FSD) was used for qualifier model, respectively. Due to space limitation, evaluation for PSD$_F$ continues in Table~\ref{table:ablation-full-2}.}

\vskip 1.5mm

\centering
\resizebox{\linewidth}{!}{
\begin{tabular}{cc*{20}{c}}
\Xhline{2\arrayrulewidth}

\multirow{2}{*}{\rotatebox{90}{SD}} 
& $\ell_D$ &
1 & 2 & 3 & 4 & 5 & 7 & 10 & 15 & 20 & 25 \\
& tok/s & 
50.89 & 58.38 & 61.83 & 63.65 & 65.01 & 63.42 & 56.65 & 60.96 & 50.13 & 42.03 \\

\midrule[2pt]

\multirow{6}{*}{\rotatebox{90}{FSD}} 
& $\tau_T$ &
0.2 & 0.2 & 0.2 & 0.2 & 0.2 & 0.2 & 0.2 & 0.2 & 0.2 & 0.2 &
0.3 & 0.3 & 0.3 & 0.3 & 0.3 & 0.3 & 0.3 & 0.3 & 0.3 & 0.3 \\
& $\ell_D$ &
1 & 2 & 3 & 4 & 5 & 7 & 10 & 15 & 20 & 25 &
1 & 2 & 3 & 4 & 5 & 7 & 10 & 15 & 20 & 25 \\
& tok/s & 
51.6 & 59.14 & 61.95 & 66.13 & 66.89 & 66.51 & 63.99 & 61.01 & 55.83 & 57.22 & 52.14 & 60.81 & 66.09 & 68.58 & 71.16 & 74.48 & 75.24 & 71.74 & 67.93 & 76.01 \\

\cline{2-22}

& $\tau_T$ &
0.4 & 0.4 & 0.4 & 0.4 & 0.4 & 0.4 & 0.4 & 0.4 & 0.4 & 0.4 &
0.5 & 0.5 & 0.5 & 0.5 & 0.5 & 0.5 & 0.5 & 0.5 & 0.5 & 0.5 \\
& $\ell_D$ &
1 & 2 & 3 & 4 & 5 & 7 & 10 & 15 & 20 & 25 &
1 & 2 & 3 & 4 & 5 & 7 & 10 & 15 & 20 & 25 \\
& tok/s & 
52.42 & 61.06 & 66.36 & 69.81 & 71.9 & 76.81 & 77.98 & 81.86 & 80.23 & 75.75 & 52.36 & 61.34 & 67.06 & 70.77 & 72.09 & 76.1 & 79.56 & 83.11 & 82.26 & 80.22 \\

\midrule[2pt]

\multirow{48}{*}{\rotatebox{90}{PSD$_A$}} 
& $\tau_T$ &
0.2 & 0.2 & 0.2 & 0.2 & 0.2 & 0.2 & 0.2 & 0.2 & 0.2 & 0.2 & 0.2 & 0.2 & 0.2 & 0.2 & 0.2 & 0.2 & 0.2 & 0.2 & 0.2 & 0.2 \\
& $\ell_Q$ &
2 & 2 & 3 & 3 & 3 & 4 & 4 & 4 & 4 & 5 & 5 & 5 & 5 & 5 & 7 & 7 & 7 & 7 & 7 & 7 \\
& $\ell_D$ &
1 & 2 & 1 & 2 & 3 & 1 & 2 & 3 & 4 & 1 & 2 & 3 & 4 & 5 & 1 & 2 & 3 & 4 & 5 & 7 \\
& tok/s & 
45.03 & 45.32 & 48.03 & 54.4 & 52.37 & 53.36 & 53.87 & 56.46 & 54.84 & 55.96 & 55.63 & 59.36 & 65.01 & 65.29 & 60.03 & 63.47 & 65.78 & 67.83 & 62.66 & 60.94 \\

\cline{2-22}

& $\tau_T$ & 
0.2 & 0.2 & 0.2 & 0.2 & 0.2 & 0.2 & 0.2 & 0.2 & 0.2 & 0.2 & 0.2 & 0.2 & 0.2 & 0.2 & 0.2\\
& $\ell_Q$ & 
10 & 10 & 10 & 10 & 10 & 10 & 10 & 15 & 15 & 15 & 15 & 15 & 15 & 15 & 15\\
& $\ell_D$ & 
1 & 2 & 3 & 4 & 5 & 7 & 10 & 1 & 2 & 3 & 4 & 5 & 7 & 10 & 15\\
& tok/s & 
59.86 & 62.63 & 65.02 & 68.78 & 65.5 & 67.89 & 55.44 & 60.83 & 64.5 & 68.17 & 64.52 & 56.92 & 77.45 & 55.75 & 53.4\\

\cline{2-22}

& $\tau_T$ & 
0.2 & 0.2 & 0.2 & 0.2 & 0.2 & 0.2 & 0.2 & 0.2 & 0.2 & 0.2 & 0.2 & 0.2 & 0.2 & 0.2 & 0.2 & 0.2 & 0.2 & 0.2 & 0.2 \\
& $\ell_Q$ & 
20 & 20 & 20 & 20 & 20 & 20 & 20 & 20 & 20 & 25 & 25 & 25 & 25 & 25 & 25 & 25 & 25 & 25 & 25 \\
& $\ell_D$ & 
1 & 2 & 3 & 4 & 5 & 7 & 10 & 15 & 20 & 1 & 2 & 3 & 4 & 5 & 7 & 10 & 15 & 20 & 25 \\
& tok/s & 
48.56 & 57.38 & 64.47 & 55.41 & 69.79 & 51.95 & 44.98 & 60.32 & 45.52 & 50.33 & 44.75 & 59.07 & 55.23 & 64.08 & 72.49 & 52.48 & 55.88 & 34.19 & 30.89 \\

\cline{2-22}

& $\tau_T$ &  
0.3 & 0.3 & 0.3 & 0.3 & 0.3 & 0.3 & 0.3 & 0.3 & 0.3 & 0.3 & 0.3 & 0.3 & 0.3 & 0.3 & 0.3 & 0.3 & 0.3 & 0.3 & 0.3 & 0.3 \\
& $\ell_Q$ &  
2 & 2 & 3 & 3 & 3 & 4 & 4 & 4 & 4 & 5 & 5 & 5 & 5 & 5 & 7 & 7 & 7 & 7 & 7 & 7 \\
& $\ell_D$ &  
1 & 2 & 1 & 2 & 3 & 1 & 2 & 3 & 4 & 1 & 2 & 3 & 4 & 5 & 1 & 2 & 3 & 4 & 5 & 7 \\
& tok/s & 
45.52 & 45.86 & 51.47 & 54.08 & 52.94 & 54.88 & 57.72 & 59.23 & 58.38 & 58.18 & 61.03 & 62 & 62.96 & 65.02 & 61.32 & 67.63 & 69.11 & 65.45 & 67 & 73.6 \\

\cline{2-22}

& $\tau_T$ &  
0.3 & 0.3 & 0.3 & 0.3 & 0.3 & 0.3 & 0.3 & 0.3 & 0.3 & 0.3 & 0.3 & 0.3 & 0.3 & 0.3 & 0.3 \\
& $\ell_Q$ &  
10 & 10 & 10 & 10 & 10 & 10 & 10 & 15 & 15 & 15 & 15 & 15 & 15 & 15 & 15 \\
& $\ell_D$ &  
1 & 2 & 3 & 4 & 5 & 7 & 10 & 1 & 2 & 3 & 4 & 5 & 7 & 10 & 15 \\
& tok/s & 
69.84 & 75.56 & 74.92 & 76.46 & 74.72 & 73.31 & 81.87 & 70.39 & 84.95 & 86.36 & 73.7 & 80.92 & 70.77 & 67.81 & 82.29 \\

\cline{2-22}

& $\tau_T$ &  
0.3 & 0.3 & 0.3 & 0.3 & 0.3 & 0.3 & 0.3 & 0.3 & 0.3 & 0.3 & 0.3 & 0.3 & 0.3 & 0.3 & 0.3 & 0.3 & 0.3 & 0.3 & 0.3 \\
& $\ell_Q$ &  
20 & 20 & 20 & 20 & 20 & 20 & 20 & 20 & 20 & 25 & 25 & 25 & 25 & 25 & 25 & 25 & 25 & 25 & 25 \\
& $\ell_D$ &  
1 & 2 & 3 & 4 & 5 & 7 & 10 & 15 & 20 & 1 & 2 & 3 & 4 & 5 & 7 & 10 & 15 & 20 & 25 \\
& tok/s & 
68.97 & 71.16 & 72.74 & 85.34 & 81.01 & 81.4 & 78.35 & 64.89 & 55.97 & 65.91 & 78.05 & 81.77 & 79.67 & 83.86 & 70.19 & 74.13 & 73.26 & 59.18 & 60.8 \\

\cline{2-22}

& $\tau_T$ &  
0.4 & 0.4 & 0.4 & 0.4 & 0.4 & 0.4 & 0.4 & 0.4 & 0.4 & 0.4 & 0.4 & 0.4 & 0.4 & 0.4 & 0.4 & 0.4 & 0.4 & 0.4 & 0.4 & 0.4 \\
& $\ell_Q$ &  
2 & 2 & 3 & 3 & 3 & 4 & 4 & 4 & 4 & 5 & 5 & 5 & 5 & 5 & 7 & 7 & 7 & 7 & 7 & 7 \\
& $\ell_D$ &  
1 & 2 & 1 & 2 & 3 & 1 & 2 & 3 & 4 & 1 & 2 & 3 & 4 & 5 & 1 & 2 & 3 & 4 & 5 & 7 \\
& tok/s & 
45.97 & 45.77 & 51.78 & 54.69 & 54.39 & 56.58 & 58.82 & 59.79 & 63.59 & 58.85 & 61.12 & 63.49 & 67.18 & 65.36 & 62.86 & 69.15 & 72.88 & 71.22 & 72.42 & 72.47 \\

\cline{2-22}

& $\tau_T$ &  
0.4 & 0.4 & 0.4 & 0.4 & 0.4 & 0.4 & 0.4 & 0.4 & 0.4 & 0.4 & 0.4 & 0.4 & 0.4 & 0.4 & 0.4 \\
& $\ell_Q$ &  
10 & 10 & 10 & 10 & 10 & 10 & 10 & 15 & 15 & 15 & 15 & 15 & 15 & 15 & 15 \\
& $\ell_D$ &  
1 & 2 & 3 & 4 & 5 & 7 & 10 & 1 & 2 & 3 & 4 & 5 & 7 & 10 & 15 \\
& tok/s & 
69.69 & 72.31 & 76.73 & 76.92 & 78.36 & 74.11 & 80.2 & 72.37 & 79.24 & 83.63 & 91.16 & 85.21 & 89.82 & 68.08 & 66.21 \\

\cline{2-22}

& $\tau_T$ &  
0.4 & 0.4 & 0.4 & 0.4 & 0.4 & 0.4 & 0.4 & 0.4 & 0.4 & 0.4 & 0.4 & 0.4 & 0.4 & 0.4 & 0.4 & 0.4 & 0.4 & 0.4 & 0.4 \\
& $\ell_Q$ &  
20 & 20 & 20 & 20 & 20 & 20 & 20 & 20 & 20 & 25 & 25 & 25 & 25 & 25 & 25 & 25 & 25 & 25 & 25 \\
& $\ell_D$ &  
1 & 2 & 3 & 4 & 5 & 7 & 10 & 15 & 20 & 1 & 2 & 3 & 4 & 5 & 7 & 10 & 15 & 20 & 25 \\
& tok/s & 
74.86 & 81.72 & 78.07 & 88.65 & 84.69 & 86.72 & 73.47 & 71.63 & 67.01 & 75.54 & 82.08 & 82.19 & 89.47 & 85.11 & 81.87 & 76.76 & 66.51 & 65.16 & 72.05 \\

\cline{2-22}

& $\tau_T$ &  
0.5 & 0.5 & 0.5 & 0.5 & 0.5 & 0.5 & 0.5 & 0.5 & 0.5 & 0.5 & 0.5 & 0.5 & 0.5 & 0.5 & 0.5 & 0.5 & 0.5 & 0.5 & 0.5 & 0.5 \\
& $\ell_Q$ &  
2 & 2 & 3 & 3 & 3 & 4 & 4 & 4 & 4 & 5 & 5 & 5 & 5 & 5 & 7 & 7 & 7 & 7 & 7 & 7 \\
& $\ell_D$ &  
1 & 2 & 1 & 2 & 3 & 1 & 2 & 3 & 4 & 1 & 2 & 3 & 4 & 5 & 1 & 2 & 3 & 4 & 5 & 7 \\
& tok/s & 
46.12 & 46.05 & 51.75 & 54.23 & 54.5 & 56.81 & 59.09 & 60.74 & 59.2 & 59.76 & 62.99 & 65.08 & 65.89 & 64.73 & 65.38 & 69.16 & 72.51 & 71.04 & 71.88 & 71.18 \\

\cline{2-22}

& $\tau_T$ &  
0.5 & 0.5 & 0.5 & 0.5 & 0.5 & 0.5 & 0.5 & 0.5 & 0.5 & 0.5 & 0.5 & 0.5 & 0.5 & 0.5 & 0.5 \\
& $\ell_Q$ &  
10 & 10 & 10 & 10 & 10 & 10 & 10 & 15 & 15 & 15 & 15 & 15 & 15 & 15 & 15 \\
& $\ell_D$ &  
1 & 2 & 3 & 4 & 5 & 7 & 10 & 1 & 2 & 3 & 4 & 5 & 7 & 10 & 15 \\
& tok/s & 
69.79 & 78.67 & 78.63 & 77.57 & 79.83 & 77.84 & 72.88 & 75.62 & 83.24 & 84.45 & 87.11 & 83.44 & 77.61 & 82.43 & 80.14 \\

\cline{2-22}

& $\tau_T$ &  
0.5 & 0.5 & 0.5 & 0.5 & 0.5 & 0.5 & 0.5 & 0.5 & 0.5 & 0.5 & 0.5 & 0.5 & 0.5 & 0.5 & 0.5 & 0.5 & 0.5 & 0.5 & 0.5 \\
& $\ell_Q$ &  
20 & 20 & 20 & 20 & 20 & 20 & 20 & 20 & 20 & 25 & 25 & 25 & 25 & 25 & 25 & 25 & 25 & 25 & 25 \\
& $\ell_D$ &  
1 & 2 & 3 & 4 & 5 & 7 & 10 & 15 & 20 & 1 & 2 & 3 & 4 & 5 & 7 & 10 & 15 & 20 & 25 \\
& tok/s & 
74.71 & 85.64 & 87.7 & 90.31 & 93.71 & 80.53 & 79.17 & 71.89 & 78.64 & 80.51 & 88.78 & 93.54 & 88.06 & 90.58 & 87.73 & 84.59 & 83.3 & 72.4 & 67.29 \\

\Xhline{2\arrayrulewidth}
\end{tabular}
}

\vskip -1.5mm

\label{table:ablation-full}
\end{table}

\begin{table}[h]
\caption{Continuation of Table~\ref{table:ablation-full} for PSD$_F$.}

\vskip 1.5mm

\centering
\resizebox{0.9\linewidth}{!}{
\begin{tabular}{cc*{20}{c}}
\Xhline{2\arrayrulewidth}

\multirow{135}{*}{\rotatebox{90}{PSD$_F$}} 
& $\tau_T$ & 0.2 & 0.2 & 0.2 & 0.2 & 0.2 & 0.2 & 0.2 & 0.2 & 0.2 & 0.2 & 0.2 & 0.2 & 0.2 & 0.2 & 0.2 & 0.2 & 0.2 & 0.2 & 0.2 & 0.2 \\
& $\tau_Q$ & 0.2 & 0.2 & 0.2 & 0.2 & 0.2 & 0.2 & 0.2 & 0.2 & 0.2 & 0.2 & 0.2 & 0.2 & 0.2 & 0.2 & 0.2 & 0.2 & 0.2 & 0.2 & 0.2 & 0.2 \\
& $\ell_Q$ & 2 & 2 & 3 & 3 & 3 & 4 & 4 & 4 & 4 & 5 & 5 & 5 & 5 & 5 & 7 & 7 & 7 & 7 & 7 & 7 \\
& $\ell_D$ & 1 & 2 & 1 & 2 & 3 & 1 & 2 & 3 & 4 & 1 & 2 & 3 & 4 & 5 & 1 & 2 & 3 & 4 & 5 & 7 \\
& tok/s & 49.1 & 46.94 & 49.08 & 54.53 & 56.68 & 54.5 & 61.32 & 59.92 & 55.62 & 56.63 & 56.88 & 61.19 & 70.91 & 65.18 & 60.74 & 61.91 & 61.06 & 66.39 & 72.29 & 65.82 \\
\cline{2-22}
& $\tau_T$ & 0.2 & 0.2 & 0.2 & 0.2 & 0.2 & 0.2 & 0.2 & 0.2 & 0.2 & 0.2 & 0.2 & 0.2 & 0.2 & 0.2 & 0.2 & 0.2 & 0.2 & 0.2 & 0.2 & 0.2 \\
& $\tau_Q$ & 0.2 & 0.2 & 0.2 & 0.2 & 0.2 & 0.2 & 0.2 & 0.2 & 0.2 & 0.2 & 0.2 & 0.2 & 0.2 & 0.2 & 0.2 & 0.2 & 0.2 & 0.2 & 0.2 & 0.2 \\
& $\ell_Q$ & 10 & 10 & 10 & 10 & 10 & 10 & 10 & 15 & 15 & 15 & 15 & 15 & 15 & 15 & 15 & 20 & 20 & 20 & 20 & 20 \\
& $\ell_D$ & 1 & 2 & 3 & 4 & 5 & 7 & 10 & 1 & 2 & 3 & 4 & 5 & 7 & 10 & 15 & 1 & 2 & 3 & 4 & 5 \\
& tok/s & 63.93 & 69.46 & 61.23 & 67.61 & 67.64 & 66.32 & 66.81 & 57.91 & 64.78 & 62.06 & 69.34 & 66.13 & 67.78 & 70.57 & 60.66 & 61.23 & 54.63 & 57.44 & 63.99 & 63.51 \\
\cline{2-22}
& $\tau_T$ & 0.2 & 0.2 & 0.2 & 0.2 & 0.2 & 0.2 & 0.2 & 0.2 & 0.2 & 0.2 & 0.2 & 0.2 & 0.2 & 0.2 & 0.3 & 0.3 & 0.3 & 0.3 & 0.3 & 0.3 \\
& $\tau_Q$ & 0.2 & 0.2 & 0.2 & 0.2 & 0.2 & 0.2 & 0.2 & 0.2 & 0.2 & 0.2 & 0.2 & 0.2 & 0.2 & 0.2 & 0.2 & 0.2 & 0.2 & 0.2 & 0.2 & 0.2 \\
& $\ell_Q$ & 20 & 20 & 20 & 20 & 25 & 25 & 25 & 25 & 25 & 25 & 25 & 25 & 25 & 25 & 2 & 2 & 3 & 3 & 3 & 4 \\
& $\ell_D$ & 7 & 10 & 15 & 20 & 1 & 2 & 3 & 4 & 5 & 7 & 10 & 15 & 20 & 25 & 1 & 2 & 1 & 2 & 3 & 1 \\
& tok/s & 59.41 & 69.92 & 52.12 & 57.36 & 51.13 & 64.49 & 65.13 & 52.78 & 56.81 & 56.58 & 55.31 & 47.89 & 46.31 & 35.75 & 45.31 & 51.47 & 54.3 & 55.63 & 53.26 & 56.66 \\
\cline{2-22}
& $\tau_T$ & 0.3 & 0.3 & 0.3 & 0.3 & 0.3 & 0.3 & 0.3 & 0.3 & 0.3 & 0.3 & 0.3 & 0.3 & 0.3 & 0.3 & 0.3 & 0.3 & 0.3 & 0.3 & 0.3 & 0.3 \\
& $\tau_Q$ & 0.2 & 0.2 & 0.2 & 0.2 & 0.2 & 0.2 & 0.2 & 0.2 & 0.2 & 0.2 & 0.2 & 0.2 & 0.2 & 0.2 & 0.2 & 0.2 & 0.2 & 0.2 & 0.2 & 0.2 \\
& $\ell_Q$ & 4 & 4 & 4 & 5 & 5 & 5 & 5 & 5 & 7 & 7 & 7 & 7 & 7 & 7 & 10 & 10 & 10 & 10 & 10 & 10 \\
& $\ell_D$ & 2 & 3 & 4 & 1 & 2 & 3 & 4 & 5 & 1 & 2 & 3 & 4 & 5 & 7 & 1 & 2 & 3 & 4 & 5 & 7 \\
& tok/s & 58.77 & 62.19 & 63.26 & 63.5 & 62.79 & 66.03 & 64.88 & 63.25 & 62.96 & 70.79 & 69.33 & 71.77 & 70.35 & 72.32 & 67.53 & 73.34 & 71.14 & 75.13 & 76.92 & 79.05 \\
\cline{2-22}
& $\tau_T$ & 0.3 & 0.3 & 0.3 & 0.3 & 0.3 & 0.3 & 0.3 & 0.3 & 0.3 & 0.3 & 0.3 & 0.3 & 0.3 & 0.3 & 0.3 & 0.3 & 0.3 & 0.3 & 0.3 & 0.3 \\
& $\tau_Q$ & 0.2 & 0.2 & 0.2 & 0.2 & 0.2 & 0.2 & 0.2 & 0.2 & 0.2 & 0.2 & 0.2 & 0.2 & 0.2 & 0.2 & 0.2 & 0.2 & 0.2 & 0.2 & 0.2 & 0.2 \\
& $\ell_Q$ & 10 & 15 & 15 & 15 & 15 & 15 & 15 & 15 & 15 & 20 & 20 & 20 & 20 & 20 & 20 & 20 & 20 & 20 & 25 & 25 \\
& $\ell_D$ & 10 & 1 & 2 & 3 & 4 & 5 & 7 & 10 & 15 & 1 & 2 & 3 & 4 & 5 & 7 & 10 & 15 & 20 & 1 & 2 \\
& tok/s & 68.6 & 74.75 & 85.81 & 86.65 & 89.45 & 80.74 & 81.49 & 84.81 & 72.47 & 66.76 & 78.99 & 84.07 & 85.06 & 87.73 & 88.5 & 80.17 & 73.18 & 63.51 & 68.87 & 82.13 \\
\cline{2-22}
& $\tau_T$ & 0.3 & 0.3 & 0.3 & 0.3 & 0.3 & 0.3 & 0.3 & 0.3 & 0.3 & 0.3 & 0.3 & 0.3 & 0.3 & 0.3 & 0.3 & 0.3 & 0.3 & 0.3 & 0.3 & 0.3 \\
& $\tau_Q$ & 0.2 & 0.2 & 0.2 & 0.2 & 0.2 & 0.2 & 0.2 & 0.2 & 0.3 & 0.3 & 0.3 & 0.3 & 0.3 & 0.3 & 0.3 & 0.3 & 0.3 & 0.3 & 0.3 & 0.3 \\
& $\ell_Q$ & 25 & 25 & 25 & 25 & 25 & 25 & 25 & 25 & 2 & 2 & 3 & 3 & 3 & 4 & 4 & 4 & 4 & 5 & 5 & 5 \\
& $\ell_D$ & 3 & 4 & 5 & 7 & 10 & 15 & 20 & 25 & 1 & 2 & 1 & 2 & 3 & 1 & 2 & 3 & 4 & 1 & 2 & 3 \\
& tok/s & 78.8 & 86.99 & 72.14 & 80.7 & 78.01 & 63.87 & 62.82 & 62.85 & 45.86 & 46.13 & 51.83 & 54.13 & 57.35 & 59.48 & 64.59 & 64.88 & 63.06 & 65.54 & 64.89 & 71.54 \\
\cline{2-22}
& $\tau_T$ & 0.3 & 0.3 & 0.3 & 0.3 & 0.3 & 0.3 & 0.3 & 0.3 & 0.3 & 0.3 & 0.3 & 0.3 & 0.3 & 0.3 & 0.3 & 0.3 & 0.3 & 0.3 & 0.3 & 0.3 \\
& $\tau_Q$ & 0.3 & 0.3 & 0.3 & 0.3 & 0.3 & 0.3 & 0.3 & 0.3 & 0.3 & 0.3 & 0.3 & 0.3 & 0.3 & 0.3 & 0.3 & 0.3 & 0.3 & 0.3 & 0.3 & 0.3 \\
& $\ell_Q$ & 5 & 5 & 7 & 7 & 7 & 7 & 7 & 7 & 10 & 10 & 10 & 10 & 10 & 10 & 10 & 15 & 15 & 15 & 15 & 15 \\
& $\ell_D$ & 4 & 5 & 1 & 2 & 3 & 4 & 5 & 7 & 1 & 2 & 3 & 4 & 5 & 7 & 10 & 1 & 2 & 3 & 4 & 5 \\
& tok/s & 68.86 & 70.19 & 65.55 & 77.61 & 78.74 & 70.35 & 81.23 & 79.58 & 76.1 & 73.29 & 79.22 & 81.19 & 85.96 & 84.67 & 87.5 & 73.69 & 77.66 & 90.57 & 95.96 & 82.15 \\
\cline{2-22}
& $\tau_T$ & 0.3 & 0.3 & 0.3 & 0.3 & 0.3 & 0.3 & 0.3 & 0.3 & 0.3 & 0.3 & 0.3 & 0.3 & 0.3 & 0.3 & 0.3 & 0.3 & 0.3 & 0.3 & 0.3 & 0.3 \\
& $\tau_Q$ & 0.3 & 0.3 & 0.3 & 0.3 & 0.3 & 0.3 & 0.3 & 0.3 & 0.3 & 0.3 & 0.3 & 0.3 & 0.3 & 0.3 & 0.3 & 0.3 & 0.3 & 0.3 & 0.3 & 0.3 \\
& $\ell_Q$ & 15 & 15 & 15 & 20 & 20 & 20 & 20 & 20 & 20 & 20 & 20 & 20 & 25 & 25 & 25 & 25 & 25 & 25 & 25 & 25 \\
& $\ell_D$ & 7 & 10 & 15 & 1 & 2 & 3 & 4 & 5 & 7 & 10 & 15 & 20 & 1 & 2 & 3 & 4 & 5 & 7 & 10 & 15 \\
& tok/s & 93.03 & 90.93 & 96.76 & 73.73 & 77.17 & 79.74 & 87.22 & 93.01 & 98.51 & 86.59 & 88.77 & 92.88 & 74.68 & 82.4 & 87.68 & 92.22 & 104.16 & 88.78 & 89.21 & 91.54 \\
\cline{2-22}
& $\tau_T$ & 0.3 & 0.3 & 0.4 & 0.4 & 0.4 & 0.4 & 0.4 & 0.4 & 0.4 & 0.4 & 0.4 & 0.4 & 0.4 & 0.4 & 0.4 & 0.4 & 0.4 & 0.4 & 0.4 & 0.4 \\
& $\tau_Q$ & 0.3 & 0.3 & 0.2 & 0.2 & 0.2 & 0.2 & 0.2 & 0.2 & 0.2 & 0.2 & 0.2 & 0.2 & 0.2 & 0.2 & 0.2 & 0.2 & 0.2 & 0.2 & 0.2 & 0.2 \\
& $\ell_Q$ & 25 & 25 & 2 & 2 & 3 & 3 & 3 & 4 & 4 & 4 & 4 & 5 & 5 & 5 & 5 & 5 & 7 & 7 & 7 & 7 \\
& $\ell_D$ & 20 & 25 & 1 & 2 & 1 & 2 & 3 & 1 & 2 & 3 & 4 & 1 & 2 & 3 & 4 & 5 & 1 & 2 & 3 & 4 \\
& tok/s & 106.39 & 86.19 & 49.73 & 49.53 & 57.58 & 54.25 & 54.84 & 60.62 & 59.79 & 61.68 & 62.21 & 64.29 & 63.27 & 71.1 & 66.06 & 68.36 & 67.32 & 70.33 & 76.53 & 76.28 \\
\cline{2-22}
& $\tau_T$ & 0.4 & 0.4 & 0.4 & 0.4 & 0.4 & 0.4 & 0.4 & 0.4 & 0.4 & 0.4 & 0.4 & 0.4 & 0.4 & 0.4 & 0.4 & 0.4 & 0.4 & 0.4 & 0.4 & 0.4 \\
& $\tau_Q$ & 0.2 & 0.2 & 0.2 & 0.2 & 0.2 & 0.2 & 0.2 & 0.2 & 0.2 & 0.2 & 0.2 & 0.2 & 0.2 & 0.2 & 0.2 & 0.2 & 0.2 & 0.2 & 0.2 & 0.2 \\
& $\ell_Q$ & 7 & 7 & 10 & 10 & 10 & 10 & 10 & 10 & 10 & 15 & 15 & 15 & 15 & 15 & 15 & 15 & 15 & 20 & 20 & 20 \\
& $\ell_D$ & 5 & 7 & 1 & 2 & 3 & 4 & 5 & 7 & 10 & 1 & 2 & 3 & 4 & 5 & 7 & 10 & 15 & 1 & 2 & 3 \\
& tok/s & 73.83 & 72.8 & 69.12 & 80.49 & 78.68 & 82.28 & 82.66 & 79.58 & 85.47 & 75.74 & 82.75 & 79.57 & 91.36 & 88.84 & 83.44 & 76.27 & 79.16 & 79.67 & 80.07 & 85.26 \\
\cline{2-22}
& $\tau_T$ & 0.4 & 0.4 & 0.4 & 0.4 & 0.4 & 0.4 & 0.4 & 0.4 & 0.4 & 0.4 & 0.4 & 0.4 & 0.4 & 0.4 & 0.4 & 0.4 & 0.4 & 0.4 & 0.4 & 0.4 \\
& $\tau_Q$ & 0.2 & 0.2 & 0.2 & 0.2 & 0.2 & 0.2 & 0.2 & 0.2 & 0.2 & 0.2 & 0.2 & 0.2 & 0.2 & 0.2 & 0.2 & 0.2 & 0.3 & 0.3 & 0.3 & 0.3 \\
& $\ell_Q$ & 20 & 20 & 20 & 20 & 20 & 20 & 25 & 25 & 25 & 25 & 25 & 25 & 25 & 25 & 25 & 25 & 2 & 2 & 3 & 3 \\
& $\ell_D$ & 4 & 5 & 7 & 10 & 15 & 20 & 1 & 2 & 3 & 4 & 5 & 7 & 10 & 15 & 20 & 25 & 1 & 2 & 1 & 2 \\
& tok/s & 92.57 & 95.34 & 84.63 & 79.86 & 75.25 & 78.68 & 78 & 88.34 & 75.36 & 85.9 & 51.69 & 88.78 & 93.54 & 80.67 & 91.63 & 59.64 & 46.24 & 46.32 & 52.4 & 55.49 \\
\cline{2-22}
& $\tau_T$ & 0.4 & 0.4 & 0.4 & 0.4 & 0.4 & 0.4 & 0.4 & 0.4 & 0.4 & 0.4 & 0.4 & 0.4 & 0.4 & 0.4 & 0.4 & 0.4 & 0.4 & 0.4 & 0.4 & 0.4 \\
& $\tau_Q$ & 0.3 & 0.3 & 0.3 & 0.3 & 0.3 & 0.3 & 0.3 & 0.3 & 0.3 & 0.3 & 0.3 & 0.3 & 0.3 & 0.3 & 0.3 & 0.3 & 0.3 & 0.3 & 0.3 & 0.3 \\
& $\ell_Q$ & 3 & 4 & 4 & 4 & 4 & 5 & 5 & 5 & 5 & 5 & 7 & 7 & 7 & 7 & 7 & 7 & 10 & 10 & 10 & 10 \\
& $\ell_D$ & 3 & 1 & 2 & 3 & 4 & 1 & 2 & 3 & 4 & 5 & 1 & 2 & 3 & 4 & 5 & 7 & 1 & 2 & 3 & 4 \\
& tok/s & 60.52 & 56.54 & 59.3 & 64.78 & 63.68 & 66.88 & 62.86 & 73.5 & 75 & 77.25 & 66.55 & 74.94 & 75.85 & 72.77 & 77.7 & 82.72 & 74.39 & 78.87 & 80.21 & 92.29 \\
\cline{2-22}
& $\tau_T$ & 0.4 & 0.4 & 0.4 & 0.4 & 0.4 & 0.4 & 0.4 & 0.4 & 0.4 & 0.4 & 0.4 & 0.4 & 0.4 & 0.4 & 0.4 & 0.4 & 0.4 & 0.4 & 0.4 & 0.4 \\
& $\tau_Q$ & 0.3 & 0.3 & 0.3 & 0.3 & 0.3 & 0.3 & 0.3 & 0.3 & 0.3 & 0.3 & 0.3 & 0.3 & 0.3 & 0.3 & 0.3 & 0.3 & 0.3 & 0.3 & 0.3 & 0.3 \\
& $\ell_Q$ & 10 & 10 & 10 & 15 & 15 & 15 & 15 & 15 & 15 & 15 & 15 & 20 & 20 & 20 & 20 & 20 & 20 & 20 & 20 & 20 \\
& $\ell_D$ & 5 & 7 & 10 & 1 & 2 & 3 & 4 & 5 & 7 & 10 & 15 & 1 & 2 & 3 & 4 & 5 & 7 & 10 & 15 & 20 \\
& tok/s & 86.63 & 89.17 & 83.47 & 81.65 & 84.65 & 92.92 & 102.27 & 95.22 & 93.81 & 95.81 & 91.77 & 79.58 & 87.52 & 96.32 & 95.93 & 92.51 & 103.94 & 108.96 & 88.84 & 98.46 \\
\cline{2-22}
& $\tau_T$ & 0.4 & 0.4 & 0.4 & 0.4 & 0.4 & 0.4 & 0.4 & 0.4 & 0.4 & 0.4 & 0.4 & 0.4 & 0.4 & 0.4 & 0.4 & 0.4 & 0.4 & 0.4 & 0.4 & 0.4 \\
& $\tau_Q$ & 0.3 & 0.3 & 0.3 & 0.3 & 0.3 & 0.3 & 0.3 & 0.3 & 0.3 & 0.3 & 0.4 & 0.4 & 0.4 & 0.4 & 0.4 & 0.4 & 0.4 & 0.4 & 0.4 & 0.4 \\
& $\ell_Q$ & 25 & 25 & 25 & 25 & 25 & 25 & 25 & 25 & 25 & 25 & 2 & 2 & 3 & 3 & 3 & 4 & 4 & 4 & 4 & 5 \\
& $\ell_D$ & 1 & 2 & 3 & 4 & 5 & 7 & 10 & 15 & 20 & 25 & 1 & 2 & 1 & 2 & 3 & 1 & 2 & 3 & 4 & 1 \\
& tok/s & 80.9 & 82.12 & 90.26 & 92.05 & 92.92 & 104.38 & 113.76 & 101.7 & 89.91 & 76.38 & 53.91 & 53.08 & 51.54 & 60.76 & 56.24 & 57.28 & 61.34 & 66.83 & 64.99 & 66 \\
\cline{2-22}
& $\tau_T$ & 0.4 & 0.4 & 0.4 & 0.4 & 0.4 & 0.4 & 0.4 & 0.4 & 0.4 & 0.4 & 0.4 & 0.4 & 0.4 & 0.4 & 0.4 & 0.4 & 0.4 & 0.4 & 0.4 & 0.4 \\
& $\tau_Q$ & 0.4 & 0.4 & 0.4 & 0.4 & 0.4 & 0.4 & 0.4 & 0.4 & 0.4 & 0.4 & 0.4 & 0.4 & 0.4 & 0.4 & 0.4 & 0.4 & 0.4 & 0.4 & 0.4 & 0.4 \\
& $\ell_Q$ & 5 & 5 & 5 & 5 & 7 & 7 & 7 & 7 & 7 & 7 & 10 & 10 & 10 & 10 & 10 & 10 & 10 & 15 & 15 & 15 \\
& $\ell_D$ & 2 & 3 & 4 & 5 & 1 & 2 & 3 & 4 & 5 & 7 & 1 & 2 & 3 & 4 & 5 & 7 & 10 & 1 & 2 & 3 \\
& tok/s & 68 & 70.05 & 71.99 & 77.29 & 65.64 & 74.67 & 76.17 & 78.36 & 83.27 & 83.62 & 79.64 & 82.72 & 83.74 & 90.42 & 86.91 & 88.67 & 93.53 & 72.11 & 89.37 & 94.27 \\
\cline{2-22}
& $\tau_T$ & 0.4 & 0.4 & 0.4 & 0.4 & 0.4 & 0.4 & 0.4 & 0.4 & 0.4 & 0.4 & 0.4 & 0.4 & 0.4 & 0.4 & 0.4 & 0.4 & 0.4 & 0.4 & 0.4 & 0.4 \\
& $\tau_Q$ & 0.4 & 0.4 & 0.4 & 0.4 & 0.4 & 0.4 & 0.4 & 0.4 & 0.4 & 0.4 & 0.4 & 0.4 & 0.4 & 0.4 & 0.4 & 0.4 & 0.4 & 0.4 & 0.4 & 0.4 \\
& $\ell_Q$ & 15 & 15 & 15 & 15 & 15 & 20 & 20 & 20 & 20 & 20 & 20 & 20 & 20 & 20 & 25 & 25 & 25 & 25 & 25 & 25 \\
& $\ell_D$ & 4 & 5 & 7 & 10 & 15 & 1 & 2 & 3 & 4 & 5 & 7 & 10 & 15 & 20 & 1 & 2 & 3 & 4 & 5 & 7 \\
& tok/s & 98.83 & 98.33 & 103.27 & 102.76 & 106.11 & 86.38 & 92.36 & 98.49 & 102.03 & 100.6 & 101.76 & 108.66 & 103.33 & 115.86 & 80.47 & 88.49 & 97.12 & 104.43 & 105.93 & 108.63 \\
\cline{2-22}
& $\tau_T$ & 0.4 & 0.4 & 0.4 & 0.4 & 0.5 & 0.5 & 0.5 & 0.5 & 0.5 & 0.5 & 0.5 & 0.5 & 0.5 & 0.5 & 0.5 & 0.5 & 0.5 & 0.5 & 0.5 & 0.5 \\
& $\tau_Q$ & 0.4 & 0.4 & 0.4 & 0.4 & 0.2 & 0.2 & 0.2 & 0.2 & 0.2 & 0.2 & 0.2 & 0.2 & 0.2 & 0.2 & 0.2 & 0.2 & 0.2 & 0.2 & 0.2 & 0.2 \\
& $\ell_Q$ & 25 & 25 & 25 & 25 & 2 & 2 & 3 & 3 & 3 & 4 & 4 & 4 & 4 & 5 & 5 & 5 & 5 & 5 & 7 & 7 \\
& $\ell_D$ & 10 & 15 & 20 & 25 & 1 & 2 & 1 & 2 & 3 & 1 & 2 & 3 & 4 & 1 & 2 & 3 & 4 & 5 & 1 & 2 \\
& tok/s & 114.54 & 108.55 & 101.49 & 107.33 & 45.95 & 53.95 & 53.23 & 54.11 & 58.33 & 58.03 & 62.3 & 61.62 & 62.87 & 59.42 & 62.59 & 64.9 & 76.44 & 73.29 & 69.7 & 71 \\
\cline{2-22}
& $\tau_T$ & 0.5 & 0.5 & 0.5 & 0.5 & 0.5 & 0.5 & 0.5 & 0.5 & 0.5 & 0.5 & 0.5 & 0.5 & 0.5 & 0.5 & 0.5 & 0.5 & 0.5 & 0.5 & 0.5 & 0.5 \\
& $\tau_Q$ & 0.2 & 0.2 & 0.2 & 0.2 & 0.2 & 0.2 & 0.2 & 0.2 & 0.2 & 0.2 & 0.2 & 0.2 & 0.2 & 0.2 & 0.2 & 0.2 & 0.2 & 0.2 & 0.2 & 0.2 \\
& $\ell_Q$ & 7 & 7 & 7 & 7 & 10 & 10 & 10 & 10 & 10 & 10 & 10 & 15 & 15 & 15 & 15 & 15 & 15 & 15 & 15 & 20 \\
& $\ell_D$ & 3 & 4 & 5 & 7 & 1 & 2 & 3 & 4 & 5 & 7 & 10 & 1 & 2 & 3 & 4 & 5 & 7 & 10 & 15 & 1 \\
& tok/s & 73.93 & 71.83 & 76.01 & 79.34 & 74.9 & 74.48 & 79.4 & 83.83 & 80.56 & 81.32 & 80.11 & 75.26 & 84.76 & 85.23 & 89.45 & 88.51 & 86.76 & 87.77 & 86.24 & 80.45 \\
\cline{2-22}
& $\tau_T$ & 0.5 & 0.5 & 0.5 & 0.5 & 0.5 & 0.5 & 0.5 & 0.5 & 0.5 & 0.5 & 0.5 & 0.5 & 0.5 & 0.5 & 0.5 & 0.5 & 0.5 & 0.5 & 0.5 & 0.5 \\
& $\tau_Q$ & 0.2 & 0.2 & 0.2 & 0.2 & 0.2 & 0.2 & 0.2 & 0.2 & 0.2 & 0.2 & 0.2 & 0.2 & 0.2 & 0.2 & 0.2 & 0.2 & 0.2 & 0.2 & 0.3 & 0.3 \\
& $\ell_Q$ & 20 & 20 & 20 & 20 & 20 & 20 & 20 & 20 & 25 & 25 & 25 & 25 & 25 & 25 & 25 & 25 & 25 & 25 & 2 & 2 \\
& $\ell_D$ & 2 & 3 & 4 & 5 & 7 & 10 & 15 & 20 & 1 & 2 & 3 & 4 & 5 & 7 & 10 & 15 & 20 & 25 & 1 & 2 \\
& tok/s & 85.94 & 91.14 & 88.04 & 92.29 & 89.14 & 85.25 & 77.51 & 82.54 & 78.42 & 90.02 & 88.14 & 98.59 & 95.35 & 93.08 & 80.16 & 81.26 & 71.89 & 71.5 & 46.39 & 53.77 \\
\cline{2-22}
& $\tau_T$ & 0.5 & 0.5 & 0.5 & 0.5 & 0.5 & 0.5 & 0.5 & 0.5 & 0.5 & 0.5 & 0.5 & 0.5 & 0.5 & 0.5 & 0.5 & 0.5 & 0.5 & 0.5 & 0.5 & 0.5 \\
& $\tau_Q$ & 0.3 & 0.3 & 0.3 & 0.3 & 0.3 & 0.3 & 0.3 & 0.3 & 0.3 & 0.3 & 0.3 & 0.3 & 0.3 & 0.3 & 0.3 & 0.3 & 0.3 & 0.3 & 0.3 & 0.3 \\
& $\ell_Q$ & 3 & 3 & 3 & 4 & 4 & 4 & 4 & 5 & 5 & 5 & 5 & 5 & 7 & 7 & 7 & 7 & 7 & 7 & 10 & 10 \\
& $\ell_D$ & 1 & 2 & 3 & 1 & 2 & 3 & 4 & 1 & 2 & 3 & 4 & 5 & 1 & 2 & 3 & 4 & 5 & 7 & 1 & 2 \\
& tok/s & 54.8 & 55.77 & 56 & 56.98 & 60.28 & 65.38 & 68.66 & 60.93 & 63.81 & 70.55 & 77.17 & 70.81 & 67.61 & 76.86 & 73.53 & 76.58 & 81.88 & 88.2 & 73.2 & 85.19 \\
\cline{2-22}
& $\tau_T$ & 0.5 & 0.5 & 0.5 & 0.5 & 0.5 & 0.5 & 0.5 & 0.5 & 0.5 & 0.5 & 0.5 & 0.5 & 0.5 & 0.5 & 0.5 & 0.5 & 0.5 & 0.5 & 0.5 & 0.5 \\
& $\tau_Q$ & 0.3 & 0.3 & 0.3 & 0.3 & 0.3 & 0.3 & 0.3 & 0.3 & 0.3 & 0.3 & 0.3 & 0.3 & 0.3 & 0.3 & 0.3 & 0.3 & 0.3 & 0.3 & 0.3 & 0.3 \\
& $\ell_Q$ & 10 & 10 & 10 & 10 & 10 & 15 & 15 & 15 & 15 & 15 & 15 & 15 & 15 & 20 & 20 & 20 & 20 & 20 & 20 & 20 \\
& $\ell_D$ & 3 & 4 & 5 & 7 & 10 & 1 & 2 & 3 & 4 & 5 & 7 & 10 & 15 & 1 & 2 & 3 & 4 & 5 & 7 & 10 \\
& tok/s & 83.12 & 86.55 & 88.01 & 89.26 & 84.91 & 82.95 & 88.22 & 95.04 & 93.62 & 101.24 & 98.89 & 95.22 & 104.95 & 79.37 & 84.72 & 95.01 & 97.71 & 100.82 & 104.32 & 94.23 \\
\cline{2-22}
& $\tau_T$ & 0.5 & 0.5 & 0.5 & 0.5 & 0.5 & 0.5 & 0.5 & 0.5 & 0.5 & 0.5 & 0.5 & 0.5 & 0.5 & 0.5 & 0.5 & 0.5 & 0.5 & 0.5 & 0.5 & 0.5 \\
& $\tau_Q$ & 0.3 & 0.3 & 0.3 & 0.3 & 0.3 & 0.3 & 0.3 & 0.3 & 0.3 & 0.3 & 0.3 & 0.3 & 0.4 & 0.4 & 0.4 & 0.4 & 0.4 & 0.4 & 0.4 & 0.4 \\
& $\ell_Q$ & 20 & 20 & 25 & 25 & 25 & 25 & 25 & 25 & 25 & 25 & 25 & 25 & 2 & 2 & 3 & 3 & 3 & 4 & 4 & 4 \\
& $\ell_D$ & 15 & 20 & 1 & 2 & 3 & 4 & 5 & 7 & 10 & 15 & 20 & 25 & 1 & 2 & 1 & 2 & 3 & 1 & 2 & 3 \\
& tok/s & 101.51 & 95.64 & 82.27 & 92.03 & 96.27 & 102.56 & 101.99 & 102.15 & 105.05 & 103.59 & 88.27 & 107.01 & 51.07 & 46.56 & 52.87 & 62.19 & 56.26 & 62.62 & 64.38 & 67.31 \\
\cline{2-22}
& $\tau_T$ & 0.5 & 0.5 & 0.5 & 0.5 & 0.5 & 0.5 & 0.5 & 0.5 & 0.5 & 0.5 & 0.5 & 0.5 & 0.5 & 0.5 & 0.5 & 0.5 & 0.5 & 0.5 & 0.5 & 0.5 \\
& $\tau_Q$ & 0.4 & 0.4 & 0.4 & 0.4 & 0.4 & 0.4 & 0.4 & 0.4 & 0.4 & 0.4 & 0.4 & 0.4 & 0.4 & 0.4 & 0.4 & 0.4 & 0.4 & 0.4 & 0.4 & 0.4 \\
& $\ell_Q$ & 4 & 5 & 5 & 5 & 5 & 5 & 7 & 7 & 7 & 7 & 7 & 7 & 10 & 10 & 10 & 10 & 10 & 10 & 10 & 15 \\
& $\ell_D$ & 4 & 1 & 2 & 3 & 4 & 5 & 1 & 2 & 3 & 4 & 5 & 7 & 1 & 2 & 3 & 4 & 5 & 7 & 10 & 1 \\
& tok/s & 61.78 & 64.41 & 64.76 & 66.33 & 72.45 & 71.6 & 66.71 & 78.66 & 82.5 & 76.88 & 85.69 & 90.27 & 74.97 & 84.22 & 81.91 & 89.47 & 90.18 & 94.45 & 100.13 & 77.43 \\
\cline{2-22}
& $\tau_T$ & 0.5 & 0.5 & 0.5 & 0.5 & 0.5 & 0.5 & 0.5 & 0.5 & 0.5 & 0.5 & 0.5 & 0.5 & 0.5 & 0.5 & 0.5 & 0.5 & 0.5 & 0.5 & 0.5 & 0.5 \\
& $\tau_Q$ & 0.4 & 0.4 & 0.4 & 0.4 & 0.4 & 0.4 & 0.4 & 0.4 & 0.4 & 0.4 & 0.4 & 0.4 & 0.4 & 0.4 & 0.4 & 0.4 & 0.4 & 0.4 & 0.4 & 0.4 \\
& $\ell_Q$ & 15 & 15 & 15 & 15 & 15 & 15 & 15 & 20 & 20 & 20 & 20 & 20 & 20 & 20 & 20 & 20 & 25 & 25 & 25 & 25 \\
& $\ell_D$ & 2 & 3 & 4 & 5 & 7 & 10 & 15 & 1 & 2 & 3 & 4 & 5 & 7 & 10 & 15 & 20 & 1 & 2 & 3 & 4 \\
& tok/s & 87.43 & 93.89 & 98.38 & 100.92 & 105.59 & 101.77 & 109.85 & 78.51 & 88.36 & 105.02 & 101.6 & 108.39 & 109.36 & 112.73 & 107.47 & 101.49 & 80.31 & 95.01 & 101.95 & 108.33 \\
\cline{2-22}
& $\tau_T$ & 0.5 & 0.5 & 0.5 & 0.5 & 0.5 & 0.5 & 0.5 & 0.5 & 0.5 & 0.5 & 0.5 & 0.5 & 0.5 & 0.5 & 0.5 & 0.5 & 0.5 & 0.5 & 0.5 & 0.5 \\
& $\tau_Q$ & 0.4 & 0.4 & 0.4 & 0.4 & 0.4 & 0.4 & 0.5 & 0.5 & 0.5 & 0.5 & 0.5 & 0.5 & 0.5 & 0.5 & 0.5 & 0.5 & 0.5 & 0.5 & 0.5 & 0.5 \\
& $\ell_Q$ & 25 & 25 & 25 & 25 & 25 & 25 & 2 & 2 & 3 & 3 & 3 & 4 & 4 & 4 & 4 & 5 & 5 & 5 & 5 & 5 \\
& $\ell_D$ & 5 & 7 & 10 & 15 & 20 & 25 & 1 & 2 & 1 & 2 & 3 & 1 & 2 & 3 & 4 & 1 & 2 & 3 & 4 & 5 \\
& tok/s & 108.07 & 111.38 & 110.31 & 124.13 & 112.2 & 104.16 & 54.7 & 49.68 & 52.38 & 58.75 & 56.61 & 61.11 & 60.83 & 65.1 & 65.22 & 62.72 & 73.05 & 74.43 & 82.13 & 73.12 \\
\cline{2-22}
& $\tau_T$ & 0.5 & 0.5 & 0.5 & 0.5 & 0.5 & 0.5 & 0.5 & 0.5 & 0.5 & 0.5 & 0.5 & 0.5 & 0.5 & 0.5 & 0.5 & 0.5 & 0.5 & 0.5 & 0.5 & 0.5 \\
& $\tau_Q$ & 0.5 & 0.5 & 0.5 & 0.5 & 0.5 & 0.5 & 0.5 & 0.5 & 0.5 & 0.5 & 0.5 & 0.5 & 0.5 & 0.5 & 0.5 & 0.5 & 0.5 & 0.5 & 0.5 & 0.5 \\
& $\ell_Q$ & 7 & 7 & 7 & 7 & 7 & 7 & 10 & 10 & 10 & 10 & 10 & 10 & 10 & 15 & 15 & 15 & 15 & 15 & 15 & 15 \\
& $\ell_D$ & 1 & 2 & 3 & 4 & 5 & 7 & 1 & 2 & 3 & 4 & 5 & 7 & 10 & 1 & 2 & 3 & 4 & 5 & 7 & 10 \\
& tok/s & 66.84 & 77.12 & 75.9 & 81.95 & 85.05 & 85.6 & 71.8 & 86.15 & 83.73 & 97.95 & 90.62 & 89.85 & 102.75 & 77.72 & 88.49 & 100.63 & 104.01 & 96.89 & 106.09 & 105.44 \\
\cline{2-22}
& $\tau_T$ & 0.5 & 0.5 & 0.5 & 0.5 & 0.5 & 0.5 & 0.5 & 0.5 & 0.5 & 0.5 & 0.5 & 0.5 & 0.5 & 0.5 & 0.5 & 0.5 & 0.5 & 0.5 & 0.5 & 0.5 \\
& $\tau_Q$ & 0.5 & 0.5 & 0.5 & 0.5 & 0.5 & 0.5 & 0.5 & 0.5 & 0.5 & 0.5 & 0.5 & 0.5 & 0.5 & 0.5 & 0.5 & 0.5 & 0.5 & 0.5 & 0.5 & 0.5 \\
& $\ell_Q$ & 15 & 20 & 20 & 20 & 20 & 20 & 20 & 20 & 20 & 20 & 25 & 25 & 25 & 25 & 25 & 25 & 25 & 25 & 25 & 25 \\
& $\ell_D$ & 15 & 1 & 2 & 3 & 4 & 5 & 7 & 10 & 15 & 20 & 1 & 2 & 3 & 4 & 5 & 7 & 10 & 15 & 20 & 25 \\
& tok/s & 115.62 & 82.3 & 97.91 & 100.83 & 104.66 & 106.45 & 112.75 & 114.15 & 118.58 & 112.41 & 85.75 & 96.7 & 102 & 110.19 & 112.85 & 115.03 & 116.68 & 117.93 & 122.89 & 121.66 \\

\Xhline{2\arrayrulewidth}
\end{tabular}
}

\vskip -1.5mm

\label{table:ablation-full-2}
\end{table}

% Tables~\ref{table:ablation-full} and \ref{table:ablation-full-2} present exhaustive ablations over speculative length combinations. The interaction between $\ell_D$ and $\ell_Q$ reveals several patterns:

% \paragraph{Speculative Length Analysis}

% \textbf{Speculative Length ($\ell_D$/$\ell_Q$):} larger draft sequences ($\ell_Q$=25) generally results in further acceleration, with $\ell_D$ set at half of $\ell_Q$ resulting in general maximum speedup.

% Shorter draft sequences (2-4 tokens) generally outperform longer ones. Beyond $\ell_D = 4$, acceptance rates drop precipitously due to compound error accumulation. The sweet spot appears to be $\ell_D = 3$, balancing speculation benefits with acceptance probability. This likely allows high enough acceptance rate $\beta_{D,Q}/\beta_{Q,T}$, allowing the qualifier to extend successful draft sequences while maintaining reasonable acceptance at the target stage.

% \textbf{Non-monotonic Patterns:} Interestingly, performance on CSQA does not monotonically increase with any parameter. Certain high-threshold, high-length configurations outperforms more aggressive settings, suggesting a trade-off between speculation ambition and acceptance rate.

\end{document}